\documentclass[sigplan,nonacm,balance=false]{acmart}
\usepackage{xspace}
\usepackage{algorithm}
\usepackage{algpseudocode}
\usepackage{makecell}
\usepackage{multirow}
\usepackage{tikz}
\usepackage{enumitem}
\usepackage{multicol}
\AtEndPreamble{%
  \hypersetup{%
    colorlinks=true,
    linkcolor=black,
    citecolor=black,
    urlcolor=ACMDarkBlue,
  }%
}
\renewcommand\footnotetextcopyrightpermission[1]{}
\AtBeginDocument{%
  }

\copyrightyear{}
\acmYear{}
\acmDOI{}
\acmConference[]{}{}{}

\algrenewcommand\algorithmicrequire{\textbf{Input:}}
\algrenewcommand\algorithmicensure{\textbf{Output:}}

\newcommand{\sysname}{FluxMoE\xspace}

\definecolor{mypurple}{RGB}{102,0,153}

\newcommand*\circled[1]{\tikz[baseline=(char.base)]{
\node[shape=circle,draw,inner sep=0.6pt] (char) {\small {#1}};}}
\newcommand{\para}[1]{\noindent\textbf{#1}}

\setitemize{itemsep=0pt,topsep=3pt,parsep=0em}

\begin{document}

\title[]{\sysname: Decoupling Expert Residency for High-Performance MoE Serving}

\author{
    Qingxiu Liu$^{1}$, 
    Yongchao He$^{2*}$, 
    Runhan Jiang$^{2}$, 
    Zion Wang$^{3}$, 
    Bohan Zhao$^{2}$,
    Mi Zhang$^{4}$,\\
    Patrick P. C. Lee$^{1}$
}
\renewcommand{\shortauthors}{Qingxiu Liu, Yongchao He, Runhan Jiang, Zion Wang, Bohan Zhao, Mi Zhang, Patrick P. C. Lee}

\affiliation{%
  \institution{$^1$CUHK \quad $^2$Unaffiliated \quad $^3$SCITIX \quad $^4$ICT, CAS}
  \city{} 
  \country{}
}

\makeatletter
\def\@fnsymbol#1{\ensuremath{\ifcase#1\or *\or \dagger\or \ddagger\or
   \mathsection\or \mathparagraph\or \|\or **\or \dagger\dagger
   \or \ddagger\ddagger \else\@ctrerr\fi}}
\makeatother

\thanks{$^*$Corresponding author.}

\begin{abstract}
Mixture-of-Experts (MoE) models have become mainstream for scaling language
models to hundreds of billions of expert parameters. Despite sparse expert
activation, existing inference engines keep all experts GPU-resident, crowding
out the key-value cache in large-batch, long-output offline workloads.
We present \sysname, which decouples experts from physical GPU residency and
adapts their footprint to available memory through a new \emph{expert paging}
abstraction. \sysname combines PagedTensor for transparent remapping, a
bandwidth-balanced hierarchy spanning losslessly compressed GPU memory and host
DRAM, and a budget-aware residency planner. Unlike CPU-GPU co-inference and
whole-layer offloading, \sysname streams weights on demand while keeping expert
computation on GPUs.
We implement \sysname atop vLLM and evaluate it on three MoE models.
For GLM-4.5 on 8$\times$H20 GPUs, \sysname delivers up to 7.2$\times$ vLLM's
throughput and 79.0\% lower average Time-Per-Output-Token (TPOT), without
measurable model-quality loss using lossless compression.
For Mixtral-8$\times$7B-Instruct on 2$\times$L40S GPUs, where weight-resident
vLLM cannot fit, \sysname delivers 4.3$\times$ KTransformers's throughput and
29.1\% lower average TPOT.

\end{abstract}

\maketitle
\pagestyle{standardpagestyle}

\begin{sloppypar}

\section{Introduction}

Over the past year, Mixture-of-Experts (MoE) architectures have become mainstream
for scaling LLMs without proportionally increasing per-token computation.
This gain comes with unprecedented weight footprints: leading open-weight MoEs
span hundreds of billions to one trillion parameters
\cite{glm45,DeepSeekAI2025DeepSeekR1IR,Bai2025KimiKO}.
Conventional inference engines \cite{kwon23,zheng24} keep all weights
GPU-resident, crowding out the key-value (KV) cache, which retains
the prior-token attention keys and values.
Since KV demand grows with active sequence count and cached length,
reduced capacity forces the scheduler to process fewer sequences per decoding
iteration, lowering throughput.
This pressure is acute in throughput-oriented offline workloads
such as synthetic-data generation \cite{tong24}, post-training rollout
generation \cite{sheng25}, and model evaluation \cite{aggarwal25},
where large batches and long outputs increase KV demand.
Adding GPUs relieves this pressure but can reduce per-GPU efficiency.
On our H20 testbed, scaling a vLLM \cite{kwon23} deployment of
Qwen3-Next-80B-A3B-Instruct \cite{qwen3technicalreport} from two to eight GPUs
via model parallelism at batch size 256 with 4,096-token outputs raises sequence-state headroom
per GPU from 10.1\,GB to 69.8\,GB but reduces per-GPU throughput by 62.4\%.
Large batches may activate most experts within each layer
(Appendix~\ref{sec:additional-analysis}), yet the layers execute sequentially.

\emph{Must the experts of all layers remain GPU-resident simultaneously?}
Our key insight is to stream expert weights while retaining the non-expert
backbone. 
Since expert pools constitute over 95\% of the model-weight footprint
(\S\ref{subsec:parameter_volume}), this makes most weight memory eligible for reclamation;
keeping the non-expert backbone resident lets its kernels run
without weight materialization, creating an overlap window for materializing the next
expert set (\S\ref{sec:moe}).
This is feasible because, although
tokens may activate different experts within a layer, they preserve the
fixed, layer-by-layer execution order.
Thus, compute kernels access an expert's weights only while executing its
layer; otherwise, those weights occupy GPU memory without contributing to
computation. Expert weights can be managed as
\emph{streamed weights}, materialized before use and released afterward,
rather than retained as persistent GPU state by conventional engines
\cite{kwon23,zheng24}.
This leads to a new streaming execution model for MoE inference:  
\begin{equation}
    \setlength{\abovedisplayskip}{4pt}
    \setlength{\belowdisplayskip}{4pt}
    \setlength{\abovedisplayshortskip}{2pt}
    \setlength{\belowdisplayshortskip}{2pt}
    \emph{model = compute graph + streamed weights.}
\end{equation}
The compute graph is static and follows a fixed layer order,
while streaming supplies the current layer's expert weights.
We formalize this model as \emph{expert paging}, which decouples experts from
persistent GPU residency and frees GPU memory for the KV cache and other runtime
state.

Realizing expert paging poses three challenges.
(i) \emph{Transparency}: expert weights must remain accessible to existing kernels
as their physical residency changes, without kernel changes.
(ii) \emph{Timely materialization}: expert weights
must be materialized fast enough to reduce inference stalls;
streaming expert weights from host DRAM over PCIe may
not finish before the layer executes.
(iii) \emph{Adaptability}: expert residency must adapt to
changing KV-cache memory pressure without incurring excessive
materialization overhead.

We design \sysname,
a streaming execution model that augments existing
MoE inference frameworks with dynamic expert materialization
through three tightly integrated mechanisms.
(i) The \emph{PagedTensor} (\S\ref{subsec:pagedtensor}) addresses transparency by
reserving stable virtual addresses for all experts and mapping them to
dedicated or reusable physical blocks, without
modifying PyTorch operators \cite{paszke19} or Triton kernels \cite{tillet19}.
(ii) The \emph{expert memory hierarchy} (\S\ref{subsec:memory_hierarchy}) addresses timely materialization by
retaining a portion of expert weights in losslessly compressed form
in GPU memory, placing the remainder in host DRAM,
and balancing GPU decompression and PCIe
transfers according to their bandwidths.
(iii) The \emph{budget-aware residency planner} (\S\ref{subsec:planner}) selects
expert residency via scheduler state and materialization measurements under
changing KV demand.

We implement \sysname atop vLLM (v0.23.0) \cite{kwon23} and evaluate it using three
MoE models \cite{jiang24,qwen3technicalreport,glm45}.
On 8$\times$H20 GPUs, \sysname achieves up to 7.2$\times$ vLLM's throughput
for GLM-4.5 \cite{glm45} and reduces average Time-Per-Output-Token (TPOT) by 79.0\% without
measurable quality loss. On 2$\times$L40S GPUs, vLLM cannot deploy
Mixtral-8$\times$7B-Instruct \cite{jiang24}; \sysname serves it with 4.3$\times$
KTransformers's \cite{chen25} throughput and 29.1\% lower average TPOT.
We make three contributions:
\begin{itemize}[leftmargin=*]
    \item[\(\bullet\)] We introduce \emph{expert paging}, a streaming
    abstraction that decouples experts from persistent
    GPU residency and reclaims memory for the KV cache and other runtime state.

    \item[\(\bullet\)] We design \sysname with PagedTensor for transparent
    tensor remapping, a bandwidth-balanced expert hierarchy,
    and a budget-aware residency planner.

    \item[\(\bullet\)] We implement \sysname atop vLLM
    \cite{kwon23} and evaluate it on three MoE models \cite{jiang24,qwen3technicalreport,glm45}
    spanning 47B--355B parameters across two NVIDIA GPU platforms.
\end{itemize}

\section{MoE Inference Memory Characterization}
\label{sec:moe}

Under a fixed GPU-memory budget, model weights compete with the KV cache.
We first identify that expert pools dominate reclaimable weight memory,
then show when more
KV capacity raises throughput. These findings motivate expert-only
streaming and adaptive expert residency.

\subsection{Expert Weights Dominate Reclaimable Memory}
\label{subsec:parameter_volume}

MoEs scale capacity with large expert pools. Sparse activation limits
per-token computation but not memory use since conventional engines keep all
experts GPU-resident \cite{kwon23,zheng24}.

\begin{figure}[!t]
\centering
\includegraphics[height=13pt]{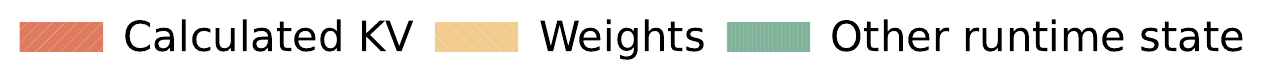}
\begin{tabular}{@{\ }c@{\ }c}
\includegraphics[width=0.43\linewidth]{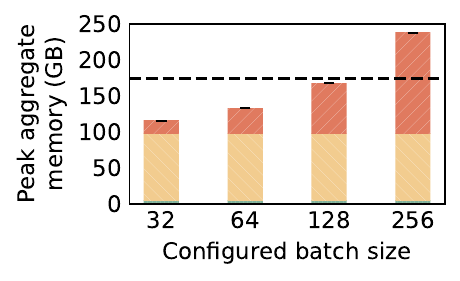}   &
\includegraphics[width=0.43\linewidth]{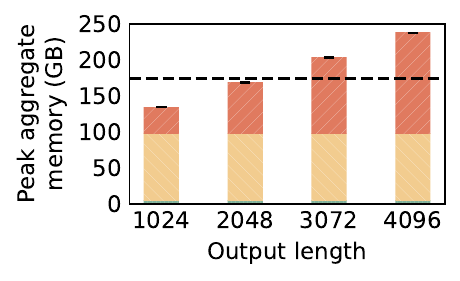}
\vspace{-3pt}\\
\makecell[c]{\small (a) Output length = 4,096} &
\makecell[c]{\small (b) Configured batch size = 256}
\end{tabular}
\vspace{-9pt}
\captionof{figure}{Calculated peak memory demand for Mixtral-8$\times$7B-Instruct
\cite{jiang24}; dashed line: 4$\times$L40S memory budget.}
\label{fig:memory_breakdown}
\vspace{-9pt}
\end{figure}

We quantify this memory pressure by running Mixtral-8$\times$7B-Instruct
\cite{jiang24} with vLLM \cite{kwon23} on 4$\times$L40S GPUs
(\S\ref{subsec:setup}). Its
weights occupy 94\,GB of the 173.9\,GB aggregate GPU budget, leaving 79.9\,GB
for the KV cache and other runtime state. We estimate peak KV
demand as total prompt-plus-output tokens times the model's BFloat16 KV bytes
per token, excluding padding. Figure~\ref{fig:memory_breakdown} shows that a
batch of 256 with 4,096-token outputs requires 139.3\,GB of KV storage,
exceeding this headroom by 59.4\,GB. At this batch size, extending the output
from 3,072 to 4,096 adds 34.4\,GB. Thus, a model may fit in GPU
memory while its target workload does not.

Expert weights constitute 95.2--96.3\% of evaluated models' parameters
(Table~\ref{tab:model_memory}), making nearly all weight memory eligible
for reclamation. Retaining the non-expert backbone avoids loading
it and lets its computation overlap expert loading. Prior MoE offloading
systems also overlap expert transfers with computation
\cite{xue24,cao25,fang25klotski}.
Whole-layer streaming would reclaim less than 5\% additional weight memory
while adding non-expert traffic and shortening the loading overlap window.
Thus, we stream only expert pools (\S\ref{sec:overview}).

\subsection{KV-Cache Capacity Constrains Throughput}
\label{subsec:kv_capacity}

We distinguish the \emph{configured batch size} $B_{\mathrm{cfg}}$, the number
of submitted sequences, from the \emph{active batch size}
$B_{\mathrm{act}}$, the number scheduled in one decoding iteration;
$\overline{B}_{\mathrm{act}}$ is its average over the run.
Each active sequence requires GPU-resident KV blocks, so
insufficient KV capacity forces vLLM to defer or preempt requests, making
$B_{\mathrm{act}}<B_{\mathrm{cfg}}$.
To isolate this effect,
we run Mixtral-8$\times$7B-Instruct \cite{jiang24} with vLLM KV offloading
on 4$\times$L40S GPUs (\S\ref{subsec:setup}) and vary only aggregate GPU KV
capacity. We report throughput and
$\overline{B}_{\mathrm{act}}$; GPU memory outside the configured KV budget
holds weights and other runtime state or remains unallocated.

\begin{table}[t]
\normalsize
\centering
\captionof{table}{Total and expert parameters of the evaluated MoEs.}
\label{tab:model_memory}
\vspace{-7pt}
\resizebox{\columnwidth}{!}{
\renewcommand\arraystretch{1.15}
\begin{tabular}{@{}ccc@{}}
\toprule[1.2pt]
\textbf{Model} & \makecell{\textbf{Total}\\\textbf{params (B)}} &
\makecell{\textbf{Expert params (B)}\\\textbf{(\% of total params)}} \\
\midrule[0.8pt]
GLM-4.5 \cite{glm45}
& 355 & 338 (95.2\%) \\
Qwen3-Next-80B-A3B-Instruct \cite{qwen3technicalreport}
& 80 & 77 (96.3\%) \\
Mixtral-8$\times$7B-Instruct \cite{jiang24}
& 47 & 45 (95.7\%) \\
\bottomrule[1.2pt]
\end{tabular}
}
\vspace{-12pt}
\end{table}

\begin{figure}[!t]
\centering
\includegraphics[height=13pt]{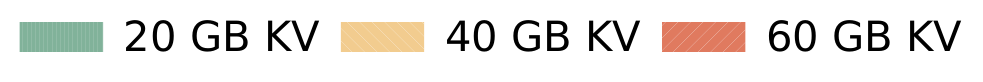}
\begin{tabular}{@{}c@{}c@{}}
\includegraphics[width=0.43\linewidth]{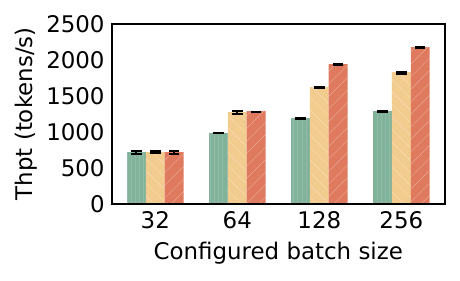} &
\includegraphics[width=0.43\linewidth]{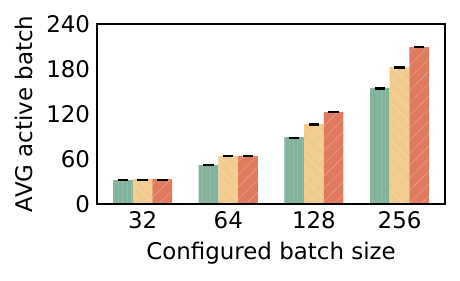}
\vspace{-3pt}\\
\makecell[c]{\small (a) Throughput} &
\makecell[c]{\small (b) Average active batch size}
\end{tabular}
\vspace{-9pt}
\captionof{figure}{Effect of aggregate GPU KV-cache capacity on
Mixtral-8$\times$7B-Instruct \cite{jiang24} at an output length of 4,096.}
\label{fig:kv_throughput}
\vspace{-12pt}
\end{figure}

Figure~\ref{fig:kv_throughput} reveals two regimes. At
$B_{\mathrm{cfg}}=32$, all configurations sustain
$\overline{B}_{\mathrm{act}}\approx B_{\mathrm{cfg}}$; increasing aggregate KV
capacity from 20\,GB to 60\,GB changes both throughput and
$\overline{B}_{\mathrm{act}}$ by less than 0.3\%. The full submitted batch
already fits, so extra KV capacity provides no benefit. At larger
$B_{\mathrm{cfg}}$, runs become KV-capacity-bound: smaller KV budgets
force the scheduler to defer or preempt more requests, lowering $\overline{B}_{\mathrm{act}}$ and
throughput. At
$B_{\mathrm{cfg}}=256$, increasing capacity from 20\,GB to 60\,GB raises
$\overline{B}_{\mathrm{act}}$ by 36.0\% and throughput by 70.0\%. As each
active sequence produces one output token per decoding iteration,
$B_{\mathrm{act}}$ determines the number of tokens produced per iteration and, together
with iteration latency, connects KV capacity to throughput.

Together, these observations imply that expert residency should account for KV
pressure. Expert pools dominate reclaimable weight memory, but their residency
reduces active batch and throughput when KV demand exceeds
capacity. Expert paging should therefore release memory under KV pressure
but retain more experts otherwise. This motivates on-demand
materialization and budget-aware residency (\S\ref{sec:overview}).

\section{Expert Paging}
\label{sec:overview}

\begin{figure}[t]
    \centering
    \setlength{\abovecaptionskip}{5pt}
    \setlength{\belowcaptionskip}{-5pt}
    \includegraphics[width=0.68\linewidth]{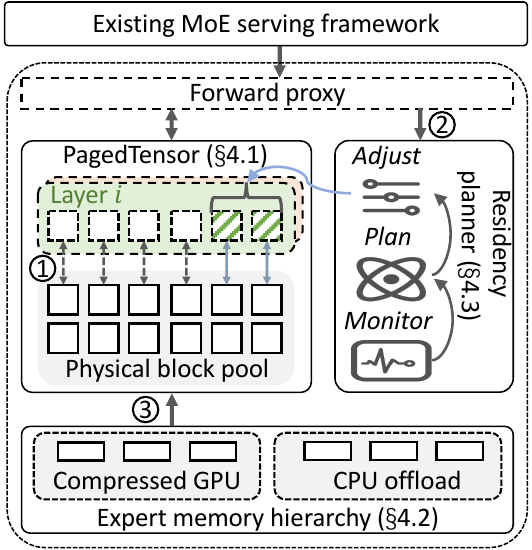}
    \caption{\sysname architecture.}
    \label{fig:architecture}
    \vspace{-9pt}
\end{figure}

Expert paging decouples experts from physical GPU residency, treating
them as virtual, on-demand resources. Rather than remaining
GPU-resident, expert weights are materialized only when their layer is scheduled,
consumed by compute kernels, and then evicted to free memory for the KV cache
and activation buffers.

We target structurally homogeneous MoEs
\cite{jiang24,qwen3technicalreport,glm45} with $N$ MoE layers in execution order, each
with $L$ routed experts of the same architecture; we leave heterogeneous layers
to future work.
Let $E^j_i$ denote routed expert $j$ in MoE layer $i$,
and $\mathcal{E}_i=\{E^1_i,E^2_i,\ldots,E^L_i\}$ denote the routed-expert set, for
$i\in\{1,\ldots,N\}$ and $j\in\{1,\ldots,L\}$. Shared experts, if present,
are excluded from $\mathcal{E}_i$. Unless otherwise stated, we use ``experts'' to 
refer to routed experts.
The model runs under model parallelism across $G$ GPU ranks, indexed by
$r\in\{1,\ldots,G\}$.
We denote $\mathcal{E}_{i,r}\subseteq\mathcal{E}_i$ as rank $r$'s expert set
for layer $i$.
Under expert parallelism (EP), all ranks divide each layer's experts evenly into disjoint, consecutive subsets.
Under tensor parallelism (TP) alone, $\mathcal{E}_{i,r}=\mathcal{E}_i$, since every rank manages a shard of every expert. By homogeneity, rank $r$ manages $L_r=|\mathcal{E}_{i,r}|$ experts per MoE layer.
Unless otherwise noted, all mechanisms in \S\ref{sec:design} operate on a
per-rank basis across the multi-GPU deployment.

During decoding, the $N$ layers execute in order within each iteration, and execution restarts from the first layer in the next iteration. A {\em forward proxy} intercepts the model's forward pass to manage expert materialization.
Before the MoE block of layer $i$ begins, the forward proxy ensures that
all rank-local experts in $\mathcal{E}_{i,r}$ are materialized and compute-ready.
After the MoE block of layer $i-1$ completes, the proxy releases
$\mathcal{E}_{i-1,r}$ and materializes $\mathcal{E}_{i+1,r}$ into the
reclaimed buffers. Since $\mathcal{E}_{i+1,r}$ is not consumed until
the MoE block of layer $i+1$, its materialization window extends
from the end of layer $i-1$'s MoE block to the beginning of layer
$i+1$'s MoE block. If materialization does not finish within this window,
layer $i+1$'s MoE block stalls. For brevity, we use layer $i$ to refer
to the entry and exit of its MoE block.
Let $R_{i,r}$ denote the rank-local expert set in this sliding window during
layer $i$'s execution.
By the materialization and
release rule above, $R_{i,r}$ satisfies:
\begin{equation}
    \setlength{\abovedisplayskip}{4pt}
    \setlength{\belowdisplayskip}{4pt}
    \setlength{\abovedisplayshortskip}{2pt}
    \setlength{\belowdisplayshortskip}{2pt}
R_{i,r}\subseteq
\mathcal{E}_{i,r}\cup\mathcal{E}_{i+1,r}.
\end{equation}

This window requires two complete rank-local MoE-layer physical blocks,
reserving $2/N$ of rank $r$'s routed-expert parameter footprint. Beyond
this window, \sysname uses any available GPU memory to retain additional experts
when KV-cache pressure is low, reducing repeated materialization.

\section{\sysname Design}
\label{sec:design}

\sysname is an MoE inference middleware that integrates with existing serving
frameworks via a lightweight forward proxy (Figure~\ref{fig:architecture}).
PagedTensor (\S\ref{subsec:pagedtensor}) maps dedicated or reusable physical blocks
to stable virtual addresses for experts, allowing kernels to access
them without modification (\circled{1}). Under KV pressure, the
{\em budget-aware residency planner} (\S\ref{subsec:planner}) selects which
layers remain uncompressed on the GPU and which expert tensors are CPU-offloaded
(\circled{2}). The expert memory hierarchy
(\S\ref{subsec:memory_hierarchy}) realizes these decisions across compressed GPU
memory and host DRAM, materializing weights into the reusable physical blocks before use
(\circled{3}).

\subsection{PagedTensor}
\label{subsec:pagedtensor}

On GPU rank $r$, PagedTensor uses CUDA virtual memory management (VMM)
\cite{cudavmm} to give rank-local expert-weight tensors stable virtual
addresses. Each layer belongs to exactly one of two residency classes:
\emph{uncompressed-resident} layers keep their weights in dedicated physical
blocks, whereas \emph{pageable} layers materialize expert weights on demand from
compressed GPU memory or host DRAM (\S\ref{subsec:memory_hierarchy}) into one
of two reusable layer blocks. Residency changes alter
only the physical backing, preserving kernel-visible addresses.
Unlike vLLM's PagedAttention \cite{kwon23}, which translates request-specific
KV-page addresses within kernels, PagedTensor exposes ordinary contiguous
tensors, requiring neither kernel changes nor in-kernel address translation.

\para{Formal model.}
Each expert has two differently sized \emph{weight tensors}: a
fused gate/up-projection tensor ($d=1$) and a down-projection tensor ($d=2$),
which PagedTensor manages separately. For type $d$ on rank $r$,
let $T_{i,r}^{(d)}$ denote layer $i$'s contiguous group of rank-local weight
tensors ordered by expert index.
For homogeneous models (\S\ref{sec:overview}), every type-$d$
tensor has byte size $\sigma_r^{(d)}$ across experts and layers; hence, the
layer-group size is
\begin{equation}
\setlength{\abovedisplayskip}{4pt}
\setlength{\belowdisplayskip}{4pt}
\setlength{\abovedisplayshortskip}{2pt}
\setlength{\belowdisplayshortskip}{2pt}
A_r^{(d)}=L_r\sigma_r^{(d)}.
\end{equation}
CUDA VMM requires allocation and mapping sizes to be aligned to its minimum
granularity $g$ \cite{cudavmm}. PagedTensor therefore defines the aligned
type-$d$ layer-group size as
\begin{equation}
S_r^{(d)}
=
\operatorname{alignUp}(A_r^{(d)},g)
=
g\left\lceil\frac{A_r^{(d)}}{g}\right\rceil .
\end{equation}

For each rank $r$ and type $d$, PagedTensor reserves
the $g$-aligned contiguous virtual region
$[v_{0,r}^{(d)},v_{0,r}^{(d)}+NS_r^{(d)})$ and partitions it
into $N$ layer ranges of size $S_r^{(d)}$.
For rank-local expert $\ell\in\{1,\ldots,L_r\}$, the type-$d$ tensor
in layer $i$ has address
\begin{equation}
\operatorname{addr}_{i,r,\ell}^{(d)}
=
v_{0,r}^{(d)}
+(i-1)S_r^{(d)}
+(\ell-1)\sigma_r^{(d)}.
\end{equation}

Given a layer-residency assignment (\S\ref{subsec:planner}),
let $\mathcal U_r$ and $\mathcal P_r$
denote rank $r$'s uncompressed-resident and pageable layers,
where
$|\mathcal U_r| + |\mathcal P_r| = N$.
For each type $d$, PagedTensor maps layer $i\in\mathcal U_r$
to a dedicated block $D_{i,r}^{(d)}$ of size $S_r^{(d)}$ and 
layer $i\in\mathcal P_r$ to one
of two reusable blocks $F_{r,b_i}^{(d)}$ of size $S_r^{(d)}$,
where $b_i=(i-1)\bmod2$.
Only residency changes alter mappings; materialization updates block contents.

\para{CUDA alignment analysis.}
Since virtual-address reservation consumes no physical GPU memory,
type-$d$ alignment overhead on rank $r$ comes from
$|\mathcal U_r|+2$ blocks, each padded from $A_r^{(d)}$ to $S_r^{(d)}$:
\begin{equation}
\setlength{\abovedisplayskip}{0pt}
\setlength{\belowdisplayskip}{4pt}
\setlength{\abovedisplayshortskip}{0pt}
\setlength{\belowdisplayshortskip}{2pt}
M_{\mathrm{pad},r}
=
\left(|\mathcal{U}_r|+2\right)
\sum_{d=1}^{2}
\left(S_r^{(d)}-A_r^{(d)}\right).
\end{equation}
All evaluated models \cite{glm45,qwen3technicalreport,jiang24} satisfy the VMM alignment requirement, so
$M_{\mathrm{pad},r}=0$.

\begin{algorithm}[t]
\caption{PagedTensor Synchronization Protocol}
\label{alg:sync_protocol}
\begingroup
\algrenewcommand\algorithmicindent{0.625em}
\begin{algorithmic}[1]
\Require GPU rank $r$, pageable layers $\mathcal{P}_r$,
tensor groups $T_{i,r}^{(d)}$, inference stream
${stream}_{r,\text{infer}}$ and loading-stream sets
${streamset}_{i,r,\text{load}}^{(d)}$

\Statex \textbf{Define} $b_i=(i-1)\bmod 2$ for each layer $i$

\Procedure{MaterializeLayer}{$i$}
    \If{$i\in\mathcal{P}_r$}
        \For{$d\in\{1,2\}$}
            \If{$last_{b_i,r}$ is defined}
                \State $\text{WaitEvent}(event_{last_{b_i,r},r}^\text{infer},
                {streamset}_{i,r,\text{load}}^{(d)})$ \Comment{\textbf{WAR}}
            \EndIf
            \State $\text{AsyncLoad}(T_{i,r}^{(d)},
            {streamset}_{i,r,\text{load}}^{(d)})$
            \State $eventset_{i,r,\text{load}}^{(d)}
            \gets\text{RecordEvents}({streamset}_{i,r,\text{load}}^{(d)})$
        \EndFor
    \EndIf
\EndProcedure

\Procedure{ForwardPass}{$i$}
    \If{$i\in\mathcal{P}_r$}
        \For{$d\in\{1,2\}$}
            \State $\text{WaitEvents}(eventset_{i,r,\text{load}}^{(d)},
            {stream}_{r,\text{infer}})$ \Comment{\textbf{RAW}}
        \EndFor
    \EndIf
    \State $\text{LaunchKernels}(i,{stream}_{r,\text{infer}})$
    \State $event_{i,r}^\text{infer}
    \gets\text{RecordEvent}({stream}_{r,\text{infer}})$
    \If{$i\in\mathcal{P}_r$}
        \State $last_{b_i,r}\gets i$ \Comment{Mark slot $b_i$ as last used by layer $i$}
    \EndIf
\EndProcedure
\end{algorithmic}
\endgroup
\end{algorithm}

\para{Async materialization.}
Overlapping reusable-block materialization with inference incurs
two ordering constraints:
\begin{itemize}[leftmargin=*]
\item \emph{Write-After-Read (WAR):} Do not overwrite a reusable block until
compute kernels reading it complete.
\item \emph{Read-After-Write (RAW):} Do not launch a pageable layer until its
weight tensors are materialized.
\end{itemize}

Algorithm~\ref{alg:sync_protocol} runs independently on each rank.
Since each reusable block backs multiple pageable layers, $last_{b_i,r}$ identifies
the layer that last used block $F_{r,b_i}^{(d)}$.
Before overwriting it, rank $r$ waits only on its local
inference-completion event (WAR), without cross-rank
synchronization. Its inference stream likewise waits for local load-completion
events before launching the layer (RAW).

\subsection{Expert Memory Hierarchy}
\label{subsec:memory_hierarchy}

Pageable layers materialize their routed-expert weight tensors
(\emph{pageable tensors}) from compressed GPU memory or host DRAM.
\sysname balances traffic between these tiers under the planner's residency
budget (\S\ref{subsec:planner}), reclaiming GPU memory for the KV cache while reducing
exposed latency.

\para{Memory hierarchy model.}
For a pageable layer on rank $r$, let $x_{\mathrm{cpu},r}$ denote the fraction
of its rank-local BFloat16 weight volume assigned to host DRAM; the remainder
uses compressed GPU memory. The backends materialize their shares concurrently
at effective bandwidths $\beta_{\mathrm{cpu},r}$ and
$\beta_{\mathrm{gpu},r}$.
As layers are homogeneous (\S\ref{sec:overview}), equalizing these
backend times yields the same ideal CPU share for every layer:
\begin{equation}
\setlength{\abovedisplayskip}{4pt}
\setlength{\belowdisplayskip}{4pt}
\setlength{\abovedisplayshortskip}{2pt}
\setlength{\belowdisplayshortskip}{2pt}
x_{\mathrm{cpu},r}^{\mathrm{bw}}
=
\frac{\beta_{\mathrm{cpu},r}}
{\beta_{\mathrm{gpu},r}+\beta_{\mathrm{cpu},r}}.
\end{equation}

This continuous target must be realized with whole tensors.
Each layer contains $L_r$ tensors of each type, and
$\sigma_r^{(1)}=2\sigma_r^{(2)}$.
Thus, down-projection tensors comprise 1/3 of its total BFloat16 weight
volume.
In our deployments, $x_{\mathrm{cpu},r}^{\mathrm{bw}}$ is well below
1/3, so down-projection tensors alone can approximate the target.
For each layer, the target CPU volume corresponds to
$3L_rx_{\mathrm{cpu},r}^{\mathrm{bw}}$ down-projection tensors.
\sysname rounds this count upward and assigns
$\lceil3L_rx_{\mathrm{cpu},r}^{\mathrm{bw}}\rceil$ such tensors to the CPU
backend. Since rounding adds less than one tensor, the resulting share
$x_{\mathrm{cpu},i,r}$ satisfies
\begin{equation}
\setlength{\abovedisplayskip}{4pt}
\setlength{\belowdisplayskip}{4pt}
\setlength{\abovedisplayshortskip}{2pt}
\setlength{\belowdisplayshortskip}{2pt}
0
\le
x_{\mathrm{cpu},i,r}-x_{\mathrm{cpu},r}^{\mathrm{bw}}
<
\frac{1}{3L_r}.
\end{equation}
Excluding budget-forced offloading, each layer's rounded CPU allocation is
within one down-projection tensor of the continuous target; fused tensors
would double this bound.

\sysname derives the backend shares from the planner's budget and the backends' relative bandwidths, rather
than scaling GPU allocation directly with model size. Thus, increasing
an MoE model's routed-expert parameter count does not require an equal increase
in GPU-resident weight storage.
GPU memory for expert weights grows with additional weights retained
uncompressed or compressed on the GPU and any enlargement of two reusable
blocks needed to hold rank-local pageable layers; offloaded weights reside in host
DRAM.

\para{Compressed GPU backend.} 
\sysname uses compressed GPU memory as its primary expert tier, reducing the
persistent footprint via on-demand decompression. Materialization is
limited by the slower of local-memory reads and decompression. Modern GPUs'
local-memory bandwidth, nearly 1\,TB/s to several TB/s, exceeds our measured
decompression throughput of approximately 400\,GB/s per GPU
(\S\ref{subsec:materialization_breakdown}); decompression
determines the materialization rate.

\sysname compresses only routed-expert weights, leaving non-expert weights
(e.g., attention projections) and shared-expert weights, if present,
uncompressed and GPU-resident. Expert weights account for over 95\% of the
evaluated models' parameter volume \cite{qwen3technicalreport,jiang24,glm45}.
Compressing only routed experts therefore captures most memory savings
while avoiding additional decompression work during the forward pass.

Our profiling finds BFloat16 expert weights exhibit the same entropy pattern as
dense-model weights: low-entropy exponent bits and nearly uniform mantissa bits
(Appendix~\ref{sec:additional-analysis}).
\sysname therefore
Huffman-encodes only the exponent bits offline and leaves the sign and mantissa
bits unchanged,
reducing routed-expert memory usage by 32.1--32.5\% across the evaluated BFloat16
models \cite{qwen3technicalreport,jiang24,glm45}
(\S\ref{subsec:materialization_breakdown}).
ZipMoE \cite{yang2026} also losslessly compresses BFloat16 exponent bits, but
targets small-batch inference on unified-memory edge devices using LZ4HC or
ZSTD with CPU decompression.

\para{CPU offload backend.} The compressed GPU backend provides high bandwidth
but is still bounded by physical GPU memory capacity. To further reduce the
GPU-resident parameter footprint, \sysname stores
expert weights in pinned host memory and transfers them over PCIe
via asynchronous DMA, following prior parameter-offloading systems \cite{sheng23,rajbhandari21}.

Modern accelerators provide tens of GB/s of PCIe bandwidth. After layer
$i-1$ completes, \sysname materializes layer $i+1$ during layer $i$'s
execution, overlapping PCIe transfer and on-device decompression with
inference. The two-layer window (\S\ref{sec:overview}) bounds in-flight
data.
PCIe alone often cannot supply expert weights fast enough to keep high-end GPUs busy.
Thus, the CPU backend acts as a complementary capacity tier, offloading weights
to expand addressable memory.

\subsection{Budget-Aware Residency Planner}
\label{subsec:planner}

On each GPU rank $r$, expert weights and KV-cache blocks compete for a fixed
local memory budget. More expert residency reduces materialization cost but
leaves less KV capacity; less residency does the reverse. \sysname's
budget-aware residency planner uses shared scheduler state and
rank-local forward and materialization measurements as empirical control
signals to select that rank's residency adjustments.

\para{Runtime control heuristic.} Let the
\emph{remaining batch size} $B_{\mathrm{rem}}$ denote the number of
unfinished sequences, including active sequences and those deferred or
preempted by the scheduler. Both $B_{\mathrm{rem}}$ and
$B_{\mathrm{act}}$ are shared across ranks and read from scheduler metadata in
every decoding iteration.

On rank $r$, let $\tau_{\mathrm{fwd},r}$ denote the forward-pass time measured
before a residency adjustment. The planner profiles it with rank-local CUDA
events, so it excludes materialization added by the candidate adjustment. We
define rank $r$'s \emph{batch-normalized effective latency} as
\begin{equation}
\setlength{\abovedisplayskip}{4pt}
\setlength{\belowdisplayskip}{4pt}
\setlength{\abovedisplayshortskip}{2pt}
\setlength{\belowdisplayshortskip}{2pt}
\tau_{\mathrm{eff},r}
=
\frac{B_{\mathrm{rem}}}{B_{\mathrm{act}}}
\tau_{\mathrm{fwd},r}.
\end{equation}
Here, $\tau_{\mathrm{eff},r}$ estimates the time for all
$B_{\mathrm{rem}}$ sequences to advance by one token.
The planner profiles materialization at layer granularity. It
measures $\tau_{\mathrm{load\_layer}}^{(i,r)}$ with rank-local CUDA events,
from load issuance until all participating local backends complete, and sets it
to zero when layer $i$ requires no materialization.
Since successive pageable layers share
the same PCIe-transfer and GPU-decompression pipelines, their loading times
accumulate over a decoding iteration:
\begin{equation}
\label{equ:loadtimeend}
\setlength{\abovedisplayskip}{2pt}
\setlength{\belowdisplayskip}{2pt}
\setlength{\abovedisplayshortskip}{2pt}
\setlength{\belowdisplayshortskip}{2pt}
\tau_{\mathrm{load},r}
=
\sum_{i=1}^{N}
\tau_{\mathrm{load\_layer}}^{(i,r)}.
\end{equation}
$\tau_{\mathrm{load},r}$ sums per-layer materialization
latencies and serves as an empirical control signal, not a measure of
stall time.
It includes latency hidden by inference; only materialization
beyond a layer's overlap window stalls its MoE block.

We combine KV-related batch pressure and aggregate materialization demand into
an empirical control signal. We define the \emph{inference-to-load ratio} as
\begin{equation}
\setlength{\abovedisplayskip}{4pt}
\setlength{\belowdisplayskip}{4pt}
\setlength{\abovedisplayshortskip}{2pt}
\setlength{\belowdisplayshortskip}{2pt}
\rho_r
=
\frac{\tau_{\mathrm{eff},r}}{\tau_{\mathrm{load},r}}.
\end{equation}
The planner uses $\rho_r$ together with $B_{\mathrm{act}}$ and
$B_{\mathrm{rem}}$ to choose a residency adjustment.
It evaluates $\rho_r$ only during decoding; it disables residency
adaptation during prefill, whose token parallelism hides most materialization.

\para{Residency control model.}
Residency reclamation proceeds in two stages.
The planner first converts all uncompressed-resident layers to pageable layers,
keeping their tensors in compressed GPU memory. Once all layers are pageable,
it enables tensor-level offloading and applies the hierarchy in
\S\ref{subsec:memory_hierarchy} to partition pageable tensors between the two
backends.
Restoration follows the reverse order. Layer-level conversion matches the
materialization pipeline, while tensor-level offload limits additional PCIe
traffic.
The rules are:
\begin{itemize}[leftmargin=*]
\item[\(\bullet\)] If $\rho_r>1$ and
$B_{\mathrm{act}}<B_{\mathrm{rem}}$, it performs the next reclamation.
\item[\(\bullet\)] If $\rho_r<\theta$ (default: 0.9), it reverses one reclamation
step.
\item[\(\bullet\)] Otherwise, residency remains unchanged.
\end{itemize}

The first rule tests reclamation only when some unfinished sequences are not
active and $\tau_{\mathrm{load},r}<\tau_{\mathrm{eff},r}$. This is an
empirical switching condition rather than a guarantee of higher throughput.
With a resizable KV allocator, reclaimed memory can expand rank $r$'s KV-cache
allocation and reduce the batch-pressure factor
$B_{\mathrm{rem}}/B_{\mathrm{act}}$. We set $\theta=0.9$, making
$[\theta,1]$ a 10\% dead zone that prevents small timing fluctuations from
repeatedly reversing residency.

\para{Runtime adaptation.} 
The planner runs only at decoding control intervals with
$B_{\mathrm{act}}>0$, ensuring that $\tau_{\mathrm{eff},r}$ is defined.
It skips intervals with no active sequences and terminates when
$B_{\mathrm{rem}}=0$ (Algorithm~\ref{alg:planner}).
The profiling (line~\ref{line:onlineprofling}) follows a producer-consumer model.
Inference and loading streams record
CUDA timestamps asynchronously, while the planner consumes only complete
measurements. Incomplete measurements are never used, and their events are
reclaimed after completion without blocking producers or adding
synchronization dependencies.

The \emph{control interval} (CI) is the number of decoding iterations between
planner runs. Within an engine launch, each run may update only rank-local
pageable-tensor backend metadata; the update takes effect at the next decoding
iteration. Layer-level residency changes require PagedTensor remapping and are
applied only between engine launches (\S\ref{subsec:planner_sensitivity}).
For tensor-level control, the \emph{adjustment granularity} (AG) defines a
step relative to one layer's $L_r$ rank-local weight tensors.
Following the fine-grained placement in
\S\ref{subsec:memory_hierarchy}, each reclamation offloads up to
$\lceil\mathrm{AG}L_r\rceil$ remaining GPU-backed down-projection tensors.
Once these are exhausted, it offloads up to
$\lceil\mathrm{AG}L_r/2\rceil$ fused gate/up-projection tensors to preserve
approximately the same BFloat16 volume. Selection favors the layer with the
smallest CPU-assigned volume, with ties broken by descending layer and
rank-local expert indices. Restoration reverses the offload sequence.
The released GPU memory is the sum of
their compressed-copy sizes. 
Once no compressed GPU copies remain, no further expert memory can be reclaimed.
We use CI=16 to balance response time and adjustment frequency and AG=0.25 to
limit the loading increase per step.

\section{Implementation}
\label{sec:impl}

We prototype \sysname as middleware atop vLLM v0.23.0 \cite{kwon23}. The
implementation comprises approximately 3.1\,K LoC of C++/CUDA for PagedTensor's
VMM allocator, stream/event coordination, and decompression kernels, and
2.1\,K LoC of Python for expert registration, backend orchestration, residency
planning, and vLLM integration.

\para{vLLM integration.}
We change 20 Python LoC across several vLLM functions:
\texttt{create\_weights} for allocating weight tensors on the CPU
to avoid their full GPU footprint,
model-specific \texttt{load\_weights} for
recording weight tensors' metadata, \texttt{load\_model} for replacing
vLLM's weight tensor storage with
PagedTensor-managed tensors at stable virtual addresses, and
\texttt{forward\_native} for hooking materialization into MoE execution. The
runtime hook waits for current-layer materialization before invoking the
unchanged MoE kernel and triggers next-layer materialization afterwards.

\para{PagedTensor.}
We implement PagedTensor using the CUDA VMM APIs \texttt{cuMemMap} and
\texttt{cuMemUnmap} \cite{cudavmm}. For each tensor type on rank $r$, it
maintains two reusable physical blocks for pageable layers and dedicated blocks for
uncompressed-resident layers (\S\ref{subsec:planner_sensitivity}).

\begin{algorithm}[t]
\caption{Budget-Aware Residency Planner}
\label{alg:planner}
\begingroup
\begin{algorithmic}[1]
\Require GPU rank $r$, threshold $\theta$ (default: $0.9$), and initial
rank-local residency state

\For{each decoding control interval with $B_{\mathrm{act}}>0$}
    \State Read shared $B_{\mathrm{rem}}$ and $B_{\mathrm{act}}$
    \label{line:onlineobserve}
    \State Read completed profiles $\tau_{\mathrm{fwd},r}$ and
    $\tau_{\mathrm{load\_layer}}^{(i,r)}$
    \label{line:onlineprofling}
    \State $\tau_{\mathrm{load},r}\gets
    \sum_{i=1}^{N}\tau_{\mathrm{load\_layer}}^{(i,r)}$
    \label{line:loadtimeaggregate}
    \State $\tau_{\mathrm{eff},r}\gets
    (B_{\mathrm{rem}}/B_{\mathrm{act}})\tau_{\mathrm{fwd},r}$
    \label{line:effective_latency}
    \State $\rho_r\gets
    \tau_{\mathrm{eff},r}/\tau_{\mathrm{load},r}$
    \Comment{$+\infty$ for zero load}
    \label{line:ratio_compute}

    \If{$\rho_r>1 \land B_{\mathrm{act}}<B_{\mathrm{rem}}$}
        \label{line:updatestart}
        \State Select the next rank-local reclamation step
        \label{line:reclaimmemory}
    \ElsIf{$\rho_r<\theta$}
        \State Select one rank-local restoration step
        \label{line:increaseresidency}
    \Else
        \State Keep rank $r$'s residency unchanged
    \EndIf
    \label{line:updateend}
\EndFor
\end{algorithmic}
\endgroup
\end{algorithm}

\para{Expert memory backends.}
We implement the compressed GPU and CPU offload backends. The GPU backend
selectively Huffman-encodes expert exponent bits \cite{huffman07} and
reconstructs them on demand with high-throughput decompression kernels
\cite{zhang25}; the CPU backend asynchronously transfers weights from pinned
host DRAM over PCIe. For each layer and rank, PagedTensor assigns each
weight-tensor type one decompression and one PCIe-transfer stream, totaling four
concurrent loading streams. Under EP,
\sysname uses vLLM's expert map to identify rank-local expert
weights without changing routing or communication.

\para{Budget-aware residency planner.}
The planner configures uncompressed GPU residency per layer and
compressed-GPU or host-DRAM backing per tensor. Within an engine launch, only
pageable-tensor backing can change, which takes effect at decoding-iteration boundaries. All
prefetches in iteration $k$ use immutable metadata version $ver_k$. For each
retired compressed GPU copy, every loading stream records a CUDA event
after its final operation that reads the copy. After issuing all iteration-$k$
prefetches, the planner atomically publishes $ver_{k+1}$ before any
iteration-$(k+1)$ prefetch, ensuring subsequent prefetches use the new
backend. The old copy remains allocated until all recorded events complete;
\sysname then releases it without device-wide synchronization.
Within a launch, memory released by tensor offloading returns to the
GPU allocator but cannot expand vLLM's KV pool, whose capacity is fixed at
initialization. Our integration fixes layer residency within each launch,
so joint expert--KV budget changes are applied only between launches
(\S\ref{subsec:planner_sensitivity}).

\section{Evaluation}
\label{sec:eval}

\subsection{Experimental Setup}
\label{subsec:setup}

\para{Testbed.}
We use two servers with different resource profiles.
The L40S server has two
24-core Intel Xeon Gold 5318Y CPUs, 2\,TB of host DRAM, and 4 NVIDIA
L40S GPUs, each with 48\,GB nominal memory (48.3\,GB reported by
\texttt{nvidia-smi}), connected via PCIe Gen4 with 32\,GB/s peak per-GPU H2D
bandwidth. The H20 server has two 48-core Intel Xeon Platinum 8468 CPUs,
2\,TB of host DRAM, and 8 NVIDIA H20 GPUs, 
each with 96\,GB nominal memory (102.6\,GB reported by
\texttt{nvidia-smi}), connected via PCIe Gen5
with 64\,GB/s peak per-GPU H2D bandwidth.

\para{Models.}
We evaluate \sysname on three MoE models.
GLM-4.5 \cite{glm45} is a production-scale MoE
whose weights exceed the aggregate memory of eight 80-GB H100 GPUs;
Mixtral-8$\times$7B-Instruct \cite{jiang24} (Mixtral for short) is widely studied; and
Qwen3-Next-80B-A3B-Instruct \cite{qwen3technicalreport} (Qwen3-Next for short) falls
between them in size.
Table~\ref{tab:eval_model_memory} summarizes the model architectures,
deployments, and memory footprints.
All deployments use vLLM's default 90\% GPU-memory budget \cite{kwon23}.
Unlike \S\ref{sec:moe}'s 4$\times$L40S setup for isolating KV-capacity
effects, we evaluate Mixtral on a constrained 2$\times$L40S
deployment, where its weights exceed vLLM's 90\% GPU-memory budget.

\para{Dataset and workloads.}
We sample input prompts from ShareGPT \cite{shareGPT}, following prior LLM
serving evaluations \cite{xiang25,jeong25,ye25,zhu25}.
Mean tokenized lengths are 44.9, 45.6, and
52.1 tokens for GLM-4.5, Qwen3-Next, and Mixtral, respectively, with a maximum
of 1,121.
Except for online-serving
evaluation, all experiments use offline serving, with all requests available
initially. We sweep either (i) $B_{\text{cfg}}\in
\{32,64,128,256\}$ with output length fixed at 4,096 or (ii) output length in
$\{1{,}024,2{,}048,3{,}072,4{,}096\}$ tokens with $B_{\text{cfg}}=256$. 
We report the average results over five runs, with error bars
as 95\% confidence intervals
under the Student's $t$-distribution.

\begin{table}[t]
\normalsize
\centering
\captionof{table}{Evaluated models and GPU memory budgets.}
\label{tab:eval_model_memory}
\vspace{-9pt}
\resizebox{0.99\columnwidth}{!}{
\renewcommand\arraystretch{1.3}
\setlength{\tabcolsep}{3pt}
\begin{tabular}{@{}cccr@{}}
\toprule[1.2pt]
\textbf{Model} & \makecell{\textbf{MoE Layers/Routed}\\\textbf{Experts per Layer}} & \textbf{GPUs} &
\makecell{\textbf{Budget $-$ Weights}\\\textbf{$=$ Headroom (GB)}} \\
\midrule[0.8pt]
GLM-4.5 \cite{glm45} & 89/160 & 8$\times$H20 &
$738.7 - 710 = 28.7$ \\
Qwen3-Next \cite{qwen3technicalreport} &
48/512 & 4$\times$L40S & $173.9 - 160 = 13.9$ \\
Mixtral \cite{jiang24} & 32/8 & 2$\times$L40S &
$86.9 - 94 = -7.1$ \\
\bottomrule[1.2pt]
\end{tabular}
}
\vspace{-9pt}
\end{table}

\para{Metrics.}
We report output throughput (tokens/s), average Time-To-First-Token
(TTFT), and average Time-Per-Output-Token (TPOT) for offline serving, and
End-to-End (E2E) latency for online serving.
TTFT, TPOT, and E2E measure submission to the first token, the mean interval
between subsequent tokens, and submission to completion, respectively.

\para{System configurations.}
All systems use a 512\,GB aggregate host-DRAM tier for KV cache offloading,
large enough to hold the full KV cache of any evaluated workload.
For a fair comparison, systems that offload expert weights to host DRAM use a
common budget of 12.5\% of routed-expert weights, the minimum needed for
Mixtral to fit in baselines.
We configure four baselines and \sysname as follows:
\begin{itemize}[leftmargin=*,nosep]
    \item[\(\bullet\)] \emph{vLLM \cite{kwon23}:} A standard weight-resident
    serving system that keeps all expert weights in GPU memory.
    \item[\(\bullet\)] \emph{vLLM-O:} A vLLM variant that tests whether
    whole-layer CPU weight offloading with prefetching suffices. It keeps
    earlier MoE layers uncompressed on the GPU, offloads the final layers
    covered by the budget to host DRAM, and prefetches layer $i+1$ while layer
    $i$ executes.
    \item[\(\bullet\)] \emph{\sysname-H:} An ablation of the
    bandwidth-balanced hierarchy (\S\ref{subsec:memory_hierarchy}) that places
    each layer entirely in one backend, using compressed GPU memory for earlier
    layers and host DRAM for the final layers covered by the budget.
    \item[\(\bullet\)] \emph{KTransformers \cite{chen25}:} A
    hybrid MoE system that distributes expert weights and
    computation across GPUs and CPUs. Rather than offloading entire layers,
    it distributes offloaded experts across all MoE layers. It uses TP
    only as it lacks EP support \cite{github}, while all
    other systems use both TP and EP. Following its configuration guide
    \cite{config}, we use 48 (96) inference
    threads on the L40S (H20) server,
    one per physical core, and two thread pools, one per NUMA node.
    \item[\(\bullet\)] \emph{\sysname:} For GLM-4.5 and Qwen3-Next, it
    keeps all routed experts in compressed GPU memory.
    As weight-resident vLLM fits on both deployments, this isolates
    compression's benefits and overheads.
    For Mixtral, we use a fixed placement to match the aggregate offloading
    budget, independently of the down-projection-first policy in \S\ref{subsec:planner}.
    Per layer, we offload one fused gate/up-projection tensor from one GPU
    and one down-projection tensor from the other (64 tensors total).
    For a fair comparison with KTransformers, we evaluate \sysname-TP, a
    TP-only variant retaining \sysname's mechanisms.
    It performs comparably to \sysname in offline and online experiments
    (Appendix~\ref{sec:additional-evaluation}).
\end{itemize}

\begin{figure*}[!t]
\centering
\includegraphics[height=15pt]{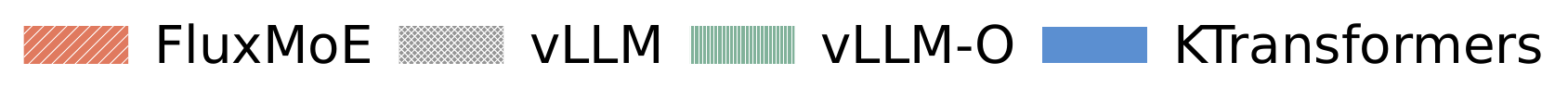}
\begin{tikzpicture}
\node[inner sep=0pt] (e2eplots) {%
\begin{tabular}{@{}c@{}c@{\hspace{4pt}}c@{}c@{}}
\includegraphics[width=0.24\linewidth]{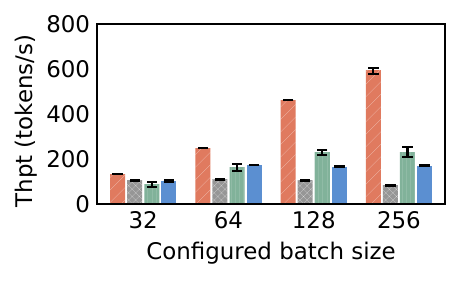} &
\includegraphics[width=0.24\linewidth]{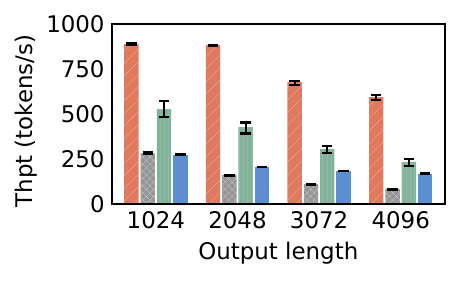} &
\includegraphics[width=0.24\linewidth]{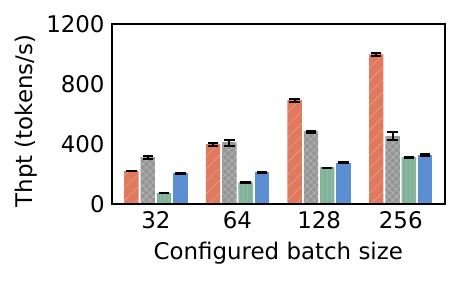} &
\includegraphics[width=0.24\linewidth]{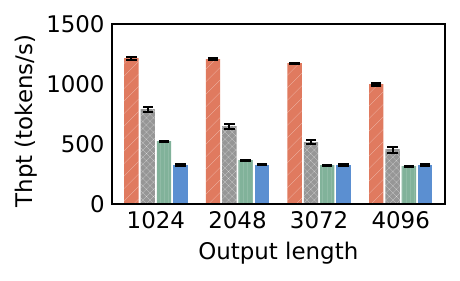} \\[-3pt]
\includegraphics[width=0.24\linewidth]{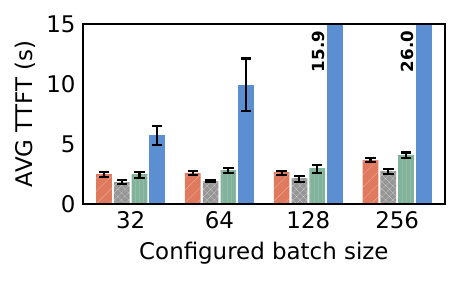} &
\includegraphics[width=0.24\linewidth]{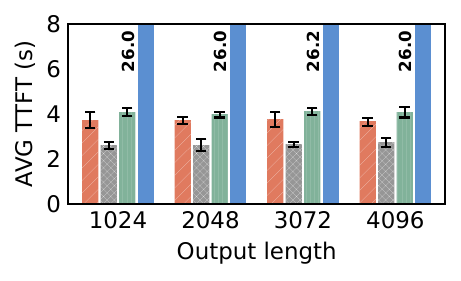} &
\includegraphics[width=0.24\linewidth]{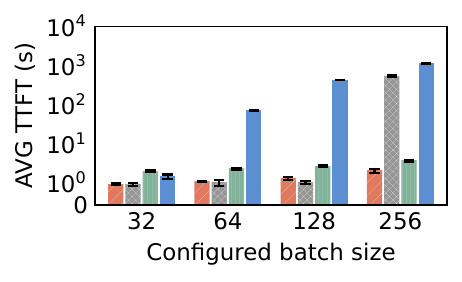} &
\includegraphics[width=0.24\linewidth]{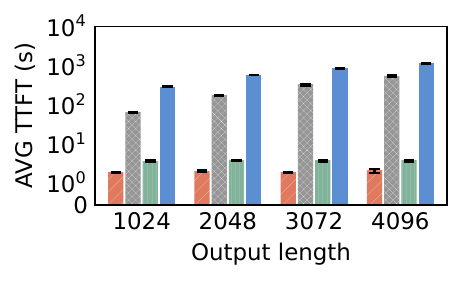} \\[-3pt]
\includegraphics[width=0.24\linewidth]{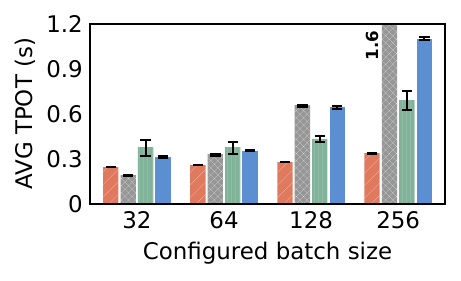} &
\includegraphics[width=0.24\linewidth]{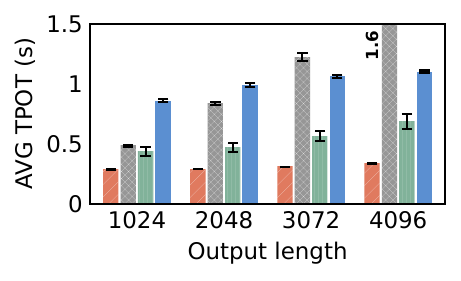} &
\includegraphics[width=0.24\linewidth]{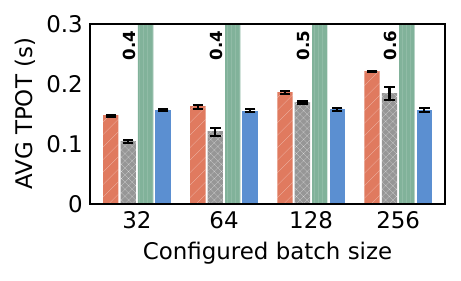} &
\includegraphics[width=0.24\linewidth]{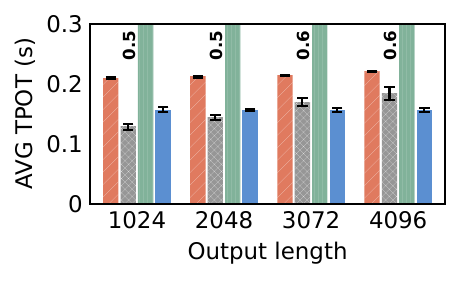} \\[-3pt]
\multicolumn{2}{c}{(a) GLM-4.5} &
\multicolumn{2}{c}{(b) Qwen3-Next}
\end{tabular}%
};
\end{tikzpicture}
\vspace{-9pt}
\captionof{figure}{(Exp\#1) Memory-sufficient regime.
For each model, left: output length = 4,096; right: $B_{\text{cfg}}$ = 256.}
\label{fig:exp1}
\vspace{-6pt}
\end{figure*}

\subsection{Overall Performance}
\label{subsec:overall}

\noindent \textbf{(Exp\#1) Memory-sufficient regime.}
Figure~\ref{fig:exp1} shows that \sysname's advantage grows with KV-cache
pressure. For GLM-4.5, at
$B_{\mathrm{cfg}}=256$, \sysname delivers 7.2$\times$, 2.6$\times$, and
3.5$\times$ the throughput of vLLM, vLLM-O, and KTransformers, respectively.
vLLM peaks at $B_{\mathrm{cfg}}=64$ and then declines as limited KV
capacity reduces $B_{\mathrm{act}}$. 
vLLM-O supports a larger $B_{\mathrm{act}}$ than vLLM through offloading but trails
\sysname since expert prefetching competes with KV-cache swapping over PCIe.
At large batches, nearly every expert receives tokens, so CPU expert
computation and output merging become KTransformers's throughput bottlenecks.

Qwen3-Next exposes the compression trade-off. At
$B_{\mathrm{cfg}}=32$, \sysname achieves 70.5\% of vLLM's throughput
since the saved memory does not raise $B_{\mathrm{act}}$; the resulting
unamortized decompression cost is expected.
At $B_{\mathrm{cfg}}=256$ and an output length of
4,096 tokens, additional KV
capacity raises $B_{\mathrm{act}}$, giving \sysname 2.2$\times$ vLLM's throughput.

\sysname also sustains throughput better as output length grows. From 1,024
to 4,096 tokens, its throughput drops by 33.5\% on GLM-4.5 and 17.8\% on
Qwen3-Next, versus 71.1\% and 42.5\% for vLLM. Its smaller expert footprint
preserves KV capacity and a higher $B_{\mathrm{act}}$ as tokens accumulate.

We next examine latencies at $B_{\mathrm{cfg}}=256$ and an output length of
4,096. On GLM-4.5, \sysname reduces TTFT and TPOT by 10.4\% and 51.3\%
versus vLLM-O, and by 86.0\% and 69.6\% versus KTransformers, respectively,
by avoiding vLLM-O's CPU-to-GPU transfers and KTransformers's CPU expert
execution and output merging.
Compared with vLLM, \sysname has 33.6\% higher TTFT but 79.0\% lower
TPOT on GLM-4.5, and 99.6\% lower TTFT but 19.9\% higher
TPOT on Qwen3-Next.
With limited KV capacity, vLLM preempts running GLM-4.5
requests, raising TPOT, but delays Qwen3-Next admission, raising TTFT.

\noindent \textbf{(Exp\#2) Memory-constrained regime.}
On Mixtral, we evaluate expert-placement strategies under the same offload
budget (Figure~\ref{fig:exp2}). At
$B_{\mathrm{cfg}}=256$ and an output length of 4,096, \sysname delivers
8.7$\times$ and 4.3$\times$ the throughput of vLLM-O and KTransformers,
respectively. It reduces TTFT and TPOT by 19.8\% and 86.9\% versus vLLM-O,
and by 99.9\% and 29.1\% versus KTransformers.
Both baselines keep GPU-resident experts uncompressed, providing less KV
capacity and a lower $B_{\mathrm{act}}$. In addition,
vLLM-O's whole-layer prefetching contends with KV-cache swapping over PCIe,
while KTransformers is limited by CPU expert execution and output merging.
\sysname increases KV
capacity and $B_{\mathrm{act}}$ through compression and reduces PCIe
contention by distributing offloaded weights across layers. Its larger
$B_{\mathrm{act}}$ raises throughput, whereas KTransformers's limited KV
capacity delays prefill, concentrating the latency gap in TTFT.

At $B_{\mathrm{cfg}}=256$ and an output length of 4,096,
\sysname improves \sysname-H's throughput by 25.3\% and
reduces its TTFT and TPOT by 28.0\% and 21.1\%, respectively.
Their equal
compressed and offloaded parameter volumes provide the same KV capacity, so
the difference comes from materialization: \sysname uses both backends per
layer, whereas \sysname-H's CPU-backed layers rely entirely on PCIe and stall
longer.

\subsection{Expert Materialization Analysis}
\label{subsec:materialization_breakdown}

\para{(Exp\#3) Lossless decompression benchmark.}
This experiment validates the compressed GPU backend by checking whether it
saves GPU memory and decompresses fast enough to avoid inference stalls.
For each model, we benchmark \sysname's selective Huffman decoder
(\S\ref{subsec:memory_hierarchy}) on one GPU of its deployment type
(Table~\ref{tab:eval_model_memory}). For a
model with $N$ MoE layers on $G$ GPUs, each of the $N\times G$ runs decodes
the routed-expert shard assigned to one layer on one GPU.
Each expert's two projection-weight tensors are decoded concurrently on separate
CUDA streams, matching inference.
We report mean $\pm$ standard deviation for latency and throughput,
defined as uncompressed BFloat16 volume divided by decoding time.

Table~\ref{tab:standalone_decompression} shows 32.1--32.5\% lossless storage
savings; Appendix~\ref{sec:additional-exp} confirms unchanged vLLM perplexity, showing
that compression does not affect model quality.
Decompression reaches 396.5--435.4\,GB/s, or
6.8--12.6$\times$ the corresponding peak PCIe H2D bandwidth,
supporting the placement of most pageable parameter volume in the compressed GPU backend
(\S\ref{subsec:memory_hierarchy}). For GLM-4.5 and Mixtral, the
2.17--3.07\,ms latency fits within Exp\#4's
5.16--21.50\,ms stall-free windows, so forwarding can hide decompression.

\begin{figure}[!t]
\centering
\includegraphics[height=13pt]{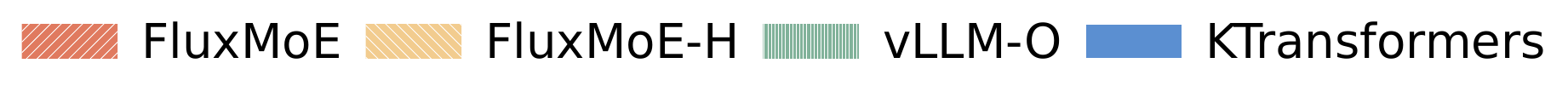}
\begin{tabular}{@{}c@{}c@{}}
\includegraphics[width=0.485\linewidth]{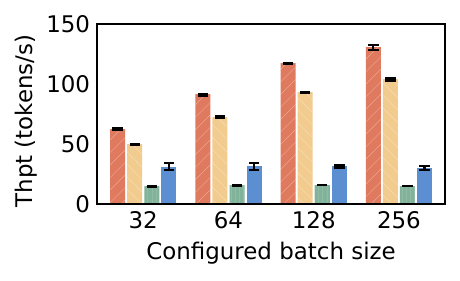} &
\includegraphics[width=0.485\linewidth]{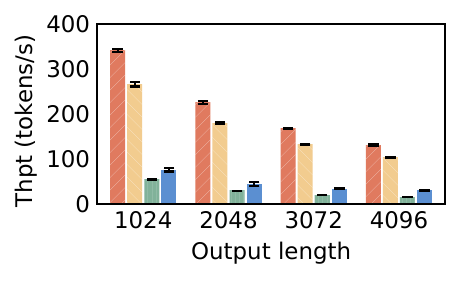} \\[-3pt]
\includegraphics[width=0.485\linewidth]{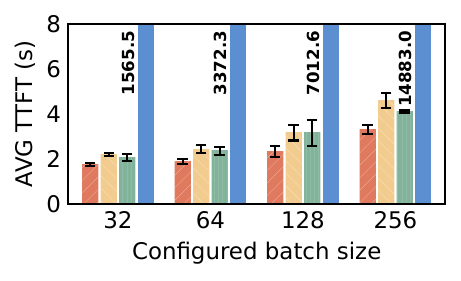} &
\includegraphics[width=0.485\linewidth]{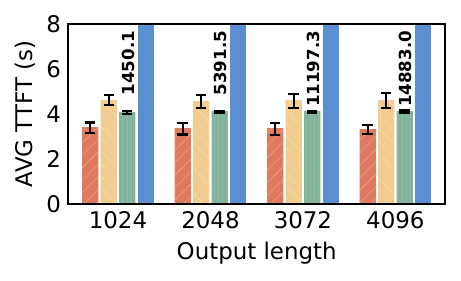} \\[-3pt]
\includegraphics[width=0.485\linewidth]{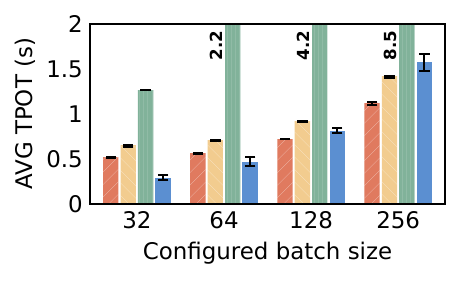} &
\includegraphics[width=0.485\linewidth]{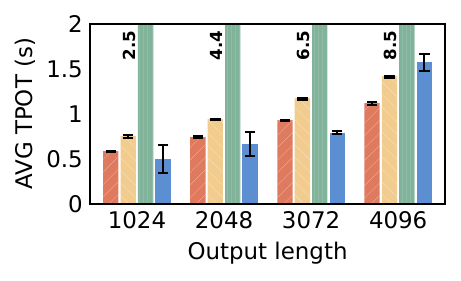} \\[-3pt]
{\small (a) Output length = 4,096} &
{\small (b) $B_{\text{cfg}}$ = 256} \\[1pt]
\end{tabular}
\vspace{-9pt}
\captionof{figure}{(Exp\#2) Memory-constrained regime for Mixtral.}
\label{fig:exp2}
\vspace{-6pt}
\end{figure}

\para{(Exp\#4) Per-layer expert-materialization breakdown.}
We measure how much materialization overlaps forwarding and whether bandwidth
balancing reduces exposed stalls under inference contention. GLM-4.5 covers
decompression-only materialization, while Mixtral covers concurrent PCIe
loading and GPU decompression, together representing both paths in
\S\ref{subsec:memory_hierarchy}.
CUDA events on the loading and inference streams measure per-layer E2E
materialization and the corresponding stall-free window; any excess is
reported as average stall across layers.
We exclude PagedTensor mapping since
it occurs only during initialization (\S\ref{subsec:pagedtensor}).

Table~\ref{tab:materialization_breakdown} shows that \sysname keeps expert
materialization largely hidden behind forwarding. For GLM-4.5, materialization
takes 2.17\,ms within a 5.16--5.82\,ms stall-free window, exposing less than
0.01\,ms of average stall per layer. For Mixtral at $B_{\mathrm{cfg}}=256$,
average stall is 33.9\% of \sysname's stall-free window, versus 51.2\% for
\sysname-H after averaging its CPU- and GPU-backed layer windows,
and \sysname reduces average per-layer stall duration
by 43.3\% compared with \sysname-H.
The reduction comes from materializing each layer through both backends
concurrently (\S\ref{subsec:memory_hierarchy}),
whereas \sysname-H's CPU-backed layers depend entirely on PCIe.
For \sysname, PCIe transfer and GPU decompression take 15.33\,ms and 3.13\,ms,
respectively, while E2E materialization takes 27.93\,ms since it also includes
queueing, stream waits, and launch delay.

\begin{figure}[!t]
\centering
\includegraphics[width=0.8\linewidth]{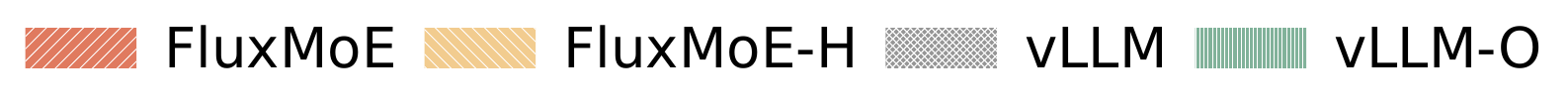}\par
\vspace{-2pt}
\begin{tabular}{@{}c@{}c@{}}
\includegraphics[width=0.485\linewidth]{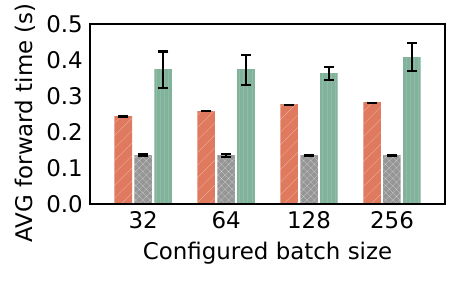} &
\includegraphics[width=0.485\linewidth]{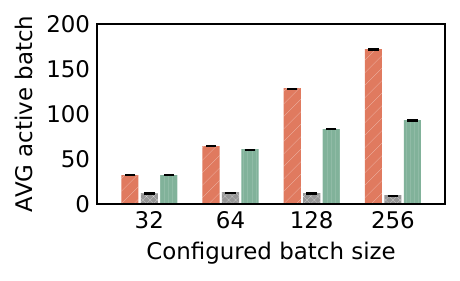} \\[-3pt]
\multicolumn{2}{c}{\small (a) GLM-4.5; Output length = 4,096} \\[1pt]
\includegraphics[width=0.485\linewidth]{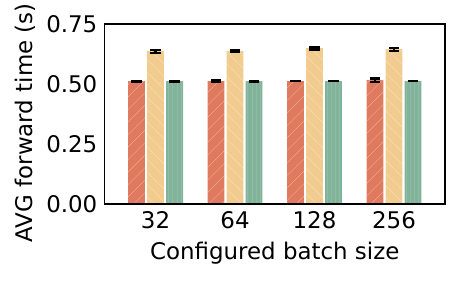} &
\includegraphics[width=0.485\linewidth]{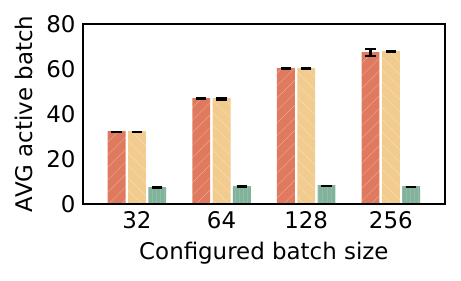} \\[-3pt]
\multicolumn{2}{c}{\small (b) Mixtral; Output length = 4,096}
\end{tabular}
\vspace{-9pt}
\captionof{figure}{(Exp\#4) Average forward time per decoding iteration (left)
and average active batch size (right).}
\label{fig:exp4}
\vspace{-9pt}
\end{figure}

\begin{table*}[!t]
\centering
\scriptsize
\captionof{table}{(Exp\#3) Lossless decompression benchmark (mean $\pm$ std.).
Uncompressed size is the BFloat16 volume decoded per layer
per GPU; for Mixtral, it covers the 87.5\% assigned to the compressed GPU
backend.}
\label{tab:standalone_decompression}
\vspace{-6pt}
\setlength{\tabcolsep}{6pt}
\renewcommand{\arraystretch}{1.05}
\resizebox{\textwidth}{!}{%
\begin{tabular}{ccccccc}
\toprule[1.2pt]
\textbf{Model} & \textbf{GPU} &
\textbf{Uncompressed size (GB)} &
\textbf{Storage saved (\%)} &
\textbf{AVG latency (ms)} & \textbf{P99 latency (ms)} &
\textbf{Thpt (GB/s)} \\
\midrule[0.8pt]
Mixtral &
L40S & 1.23 & 32.4 & $3.07\pm0.14$ & $3.13\pm0.14$ & $401.9\pm0.8$ \\
Qwen3-Next &
L40S & 0.81 & 32.1 & $2.03\pm0.01$ & $2.07\pm0.02$ & $396.5\pm2.3$ \\
GLM-4.5 &
H20 & 0.94 & 32.5 & $2.17\pm0.01$ & $2.17\pm0.01$ & $435.4\pm2.6$ \\
\bottomrule[1.2pt]
\end{tabular}
}
\vspace{-6pt}
\end{table*}

\begin{table*}[!t]
\centering
\small
\captionof{table}{(Exp\#4) Per-layer expert materialization at output length
4,096 (mean $\pm$ std., ms). CPU Offloading and Decompression report isolated
PCIe-transfer and decompression time, respectively. E2E materialization spans
load issuance to tensor readiness, including queueing, stream waits, and launch
delay; Stall-free window spans issuance until tensor use; AVG stall is the
remaining wait, averaged across layers. For \sysname-H, C and D denote CPU-
and compressed-GPU-backed layers, respectively.}
\label{tab:materialization_breakdown}
\vspace{-8pt}
\scriptsize
\setlength{\tabcolsep}{6pt}
\renewcommand{\arraystretch}{0.96}
\resizebox{0.97\textwidth}{!}{%
\begin{tabular}{@{}cccccccc@{}}
\toprule[1.2pt]
\textbf{Model} & \textbf{Method} & $\mathbf{B}_{\mathbf{cfg}}$ &
\textbf{CPU Offloading} & \textbf{Decompression} & \textbf{E2E materialization} &
\textbf{Stall-free window} & \textbf{AVG stall} \\
\midrule[0.8pt]
\multirow{2}{*}{GLM-4.5}
& \sysname & 32  & -- & $2.17\pm0.01$ & $2.17\pm0.01$ &
$5.16\pm0.43$ & $<0.01$ \\[2pt]
& \sysname & 256 & -- & $2.17\pm0.01$ & $2.17\pm0.01$ &
$5.82\pm1.15$ & $<0.01$ \\
\midrule[0.5pt]
\multirow{4}{*}[-1.5em]{Mixtral}
& \sysname   & 32  & $15.37\pm5.00$ & $3.12\pm0.15$ &
$28.13\pm1.62$ & $21.19\pm6.66$ & $7.72\pm4.97$ \\
& \sysname-H & 32  & $126.10\pm12.54$ & $3.50\pm0.01$ &
\makecell{C: $200.74\pm33.34$\\D: $3.51\pm0.01$} &
\makecell{C: $100.30\pm64.41$\\D: $13.46\pm19.00$} &
$13.05\pm36.27$ \\
\cmidrule(l){2-8}
& \sysname   & 256 & $15.33\pm4.95$ & $3.13\pm0.15$ &
$27.93\pm1.53$ & $21.50\pm6.65$ & $7.29\pm4.79$ \\
& \sysname-H & 256 & $140.58\pm12.12$ & $3.51\pm0.01$ &
\makecell{C: $204.00\pm29.46$\\D: $3.52\pm0.02$} &
\makecell{C: $105.27\pm70.27$\\D: $13.66\pm16.37$} &
$12.86\pm37.18$ \\
\bottomrule[1.2pt]
\end{tabular}%
}
\vspace{-6pt}
\end{table*}

Expert materialization remains stable as batch size grows. From
$B_{\mathrm{cfg}}=32$ to 256, GLM-4.5's E2E materialization latency stays at
2.17\,ms, while Mixtral's changes by less than 1\%, with no systematic increase
in exposed stall. Intermediate results at $B_{\mathrm{cfg}}=64$ and 128 in
Appendix~\ref{sec:additional-evaluation} support this conclusion.

Figure~\ref{fig:exp4} shows the cost-benefit trade-off behind \sysname's
throughput gain: each iteration takes longer but processes more sequences.
For GLM-4.5 at $B_{\mathrm{cfg}}=256$, compared with vLLM, \sysname increases
$\overline{B}_{\mathrm{act}}$ by 17.6$\times$ (benefit) and forward time by
1.1$\times$ (cost), due to the larger batch and GPU contention from
decompression.
For Mixtral, compared with vLLM-O, \sysname increases
$\overline{B}_{\mathrm{act}}$ by 7.8$\times$ (benefit) while increasing forward
time by only 0.8\% (cost). Compared with \sysname-H, which has the same expert
footprint, KV capacity, and nearly identical $\overline{B}_{\mathrm{act}}$,
\sysname's lower materialization stall reduces forward time by 20.0\%. This
analysis is consistent with
Figures~\ref{fig:exp1} and~\ref{fig:exp2}.

\subsection{Residency Planner Analysis}\label{subsec:planner_sensitivity}

The planner adjusts expert placement between GPU
and CPU in response to KV-cache pressure (\S\ref{subsec:planner}).
We evaluate sensitivity to the planner's control interval (CI) and adjustment
granularity (AG) (Exp\#5), then compare adaptive and fixed residency under
changing KV-cache pressure (Exp\#6).

\para{(Exp\#5) Residency planner sensitivity.}
We compare against fixed residency with all routed-expert tensors compressed
in GPU memory.
At an output length of 4,096, we use $B_{\mathrm{cfg}}=256$ for GLM-4.5 and
Qwen3-Next, but $B_{\mathrm{cfg}}=32$ for Mixtral since its no-offload
configuration leaves insufficient GPU memory for runtime state at
$B_{\mathrm{cfg}}=256$.
Before the first adjustment, the
default CI=16 corresponds to 3.0--4.5\,s across models. We sweep CI over
$\{8,16,32,64\}$ at AG=0.25 and AG over
$\{0.25,0.5,0.75,1.0\}$ at CI=16.
All runs remain in the down-projection stage: each reclamation offloads
$\lceil\mathrm{AG}L_r\rceil$ tensors per rank across pageable layers
($L_r$ at AG=1.0); no fused gate/up-projection tensor is selected.

\begin{figure}[!t]
\centering
\includegraphics[width=0.92\linewidth]{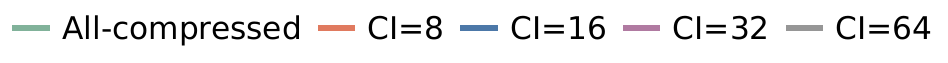}\par
\begin{tabular}{@{}c@{}c@{}c@{}}
\includegraphics[width=0.325\linewidth]{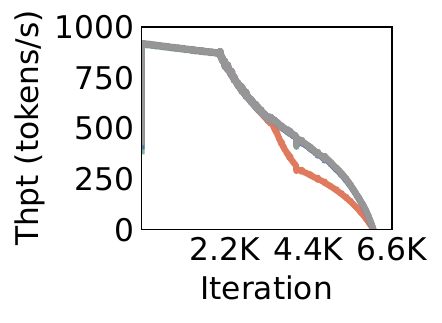} &
\includegraphics[width=0.325\linewidth]{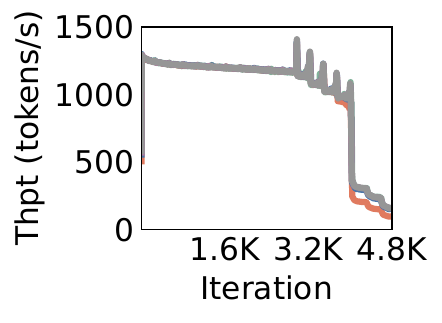} &
\includegraphics[width=0.325\linewidth]{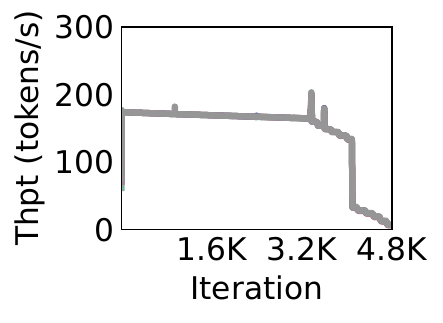} \\[-3pt]
\multicolumn{3}{c}{\small (a) AG=0.25}
\end{tabular}
\par\vspace{2pt}
\includegraphics[width=0.98\linewidth]{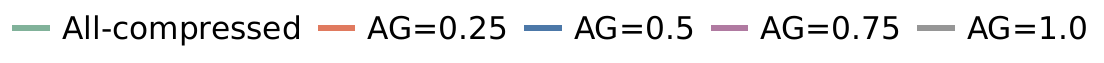}\par
\begin{tabular}{@{}c@{}c@{}c@{}}
\includegraphics[width=0.325\linewidth]{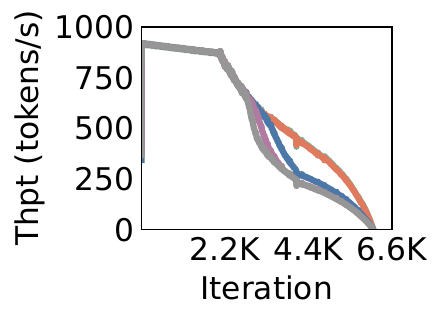} &
\includegraphics[width=0.325\linewidth]{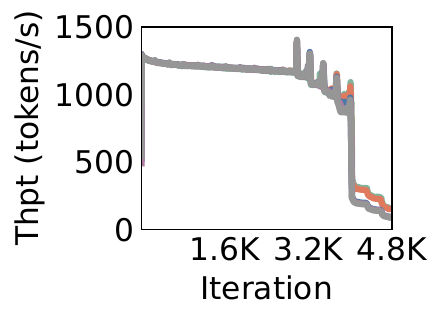} &
\includegraphics[width=0.325\linewidth]{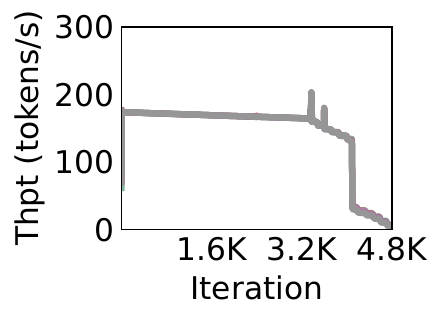} \\[-3pt]
\multicolumn{3}{c}{\small (b) CI=16}
\end{tabular}
\vspace{-12pt}
\captionof{figure}{(Exp\#5) Residency planner sensitivity. From left to right:
GLM-4.5, Qwen3-Next, and Mixtral.}
\label{fig:exp5}
\vspace{-9pt}
\end{figure}

Figure~\ref{fig:exp5} shows that the default CI=16 and AG=0.25 remain close to fixed
residency while releasing 48.8\,GB, 6.9\,GB, and 1.9\,GB of GPU memory for GLM-4.5, Qwen3-Next,
and Mixtral, respectively, equal to 10.8\%, 6.6\%, and 3.1\% of their
compressed routed-expert storage (Appendix~\ref{sec:additional-evaluation}).
H20's higher PCIe bandwidth permits
more adjustments for GLM-4.5, while Mixtral's larger tensors quickly make
loading limiting, explaining its smaller release and lower sensitivity
to CI and AG.
For GLM-4.5 and Qwen3-Next, all settings initially track fixed residency while
$B_{\mathrm{act}}=B_{\mathrm{cfg}}$. Once offloading begins, CI values of 16,
32, and 64 remain close to the baseline, whereas CI=8 releases 2.0$\times$
and 1.9$\times$ as much GPU memory as CI=16,
increasing PCIe traffic and reducing throughput.
At CI=16, larger AG lowers throughput by
offloading more tensors per adjustment; AG=0.25 remains closest to the baseline.

\para{(Exp\#6) Emulated runtime residency adaptation.}
As vLLM fixes KV capacity at initialization, we emulate adaptation
by splitting a workload with $B_{\mathrm{cfg}}=256$ and output length 4,096
across 41 engine launches. The first receives the original prompts; later
launches concatenate each prompt with its accumulated output to reconstruct
the growing KV state. Each of the first 40 produces at least
$256\times100$ new tokens in aggregate; the final launch produces the
remainder.

The first launch maximizes expert residency after reserving other runtime state
and enough KV cache to prefill one request. Before each
subsequent launch, the planner uses the preceding launch's measurements to
adjust residency: it first makes uncompressed layers pageable, potentially
several at once since each launch spans multiple control intervals,
and then offloads tensors with AG=0.25 once all layers are pageable. 
These transitions reverse as KV pressure falls.

We concatenate decoding iterations across launches. To summarize
performance, we compute \emph{aggregate throughput} as the total output tokens
divided by the sum of the initial launch's prefill time and all decoding
times, excluding reconstruction prefills and relaunch overhead.
The baseline uses \sysname's configuration from \S\ref{subsec:setup} in one launch;
replaying the configuration across 41 launches changes aggregate
throughput by less than 5\% across models.

\begin{figure}[!t]
\centering
\includegraphics[width=0.85\linewidth]{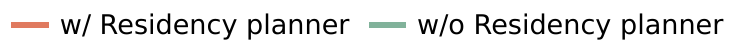}\par
\vspace{-3pt}
\begin{tabular}{@{}c@{}c@{}c@{}}
\includegraphics[width=0.325\linewidth]{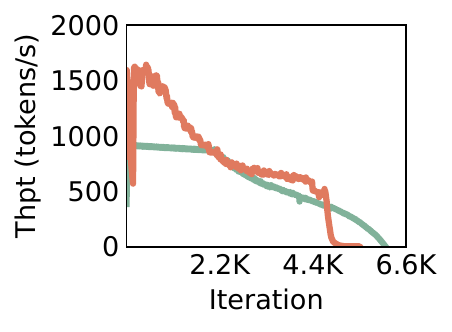} &
\includegraphics[width=0.325\linewidth]{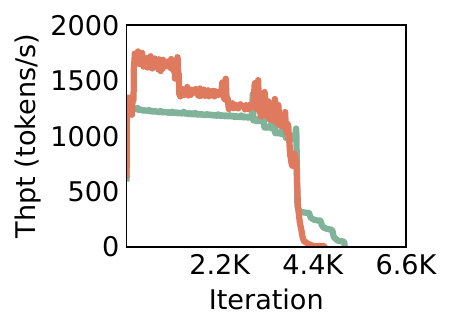} &
\includegraphics[width=0.325\linewidth]{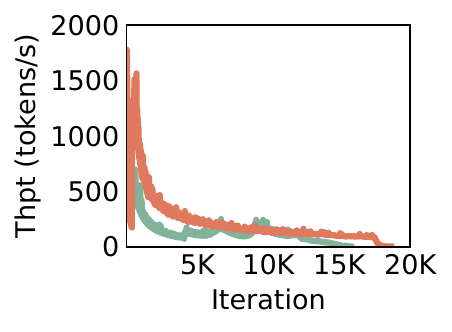} 
\end{tabular}
\vspace{-14pt}
\captionof{figure}{(Exp\#6) Emulated runtime residency adaptation. From left to right:
GLM-4.5, Qwen3-Next, and Mixtral.}
\label{fig:exp6}
\vspace{-10pt}
\end{figure}

Figure~\ref{fig:exp6} shows that the planner initially achieves higher
per-iteration throughput than fixed residency on all three models by keeping
more expert weights uncompressed and GPU-resident. As KV pressure grows, paging
and offloading expand KV capacity at the cost of loading overhead; the final
throughput drop reflects request completion.
For GLM-4.5 and Qwen3-Next, reclaimed memory reduces request suspension and
sustains a larger $\overline{B}_{\mathrm{act}}$, allowing completion in fewer
iterations. For Mixtral, the planner requires more iterations and has a smaller
$\overline{B}_{\mathrm{act}}$ than fixed residency, but maintains higher
per-iteration throughput through most of decoding.
Under this multi-launch emulation, adaptive residency improves aggregate
throughput over fixed residency by 20.9\%, 16.9\%, and 42.7\% for GLM-4.5,
Qwen3-Next, and Mixtral, respectively.
Because a dynamic-KV runtime would incur unmodeled in-run resizing and
reconfiguration overhead, these gains are 
an optimistic estimate of the potential benefit
of integrating a resizable KV allocator.

\subsection{Online Serving Performance}
\label{subsec:online_performance}

We next test whether \sysname also
reduces latency under asynchronous arrivals and varying online load.

\para{(Exp\#7) Online serving performance.}
We use the same ShareGPT prompts \cite{shareGPT} as in offline evaluation, but
keep their original response lengths.
Following prior evaluations \cite{yu22,kwon23,zhu25,gao25},
we model request arrivals as a Poisson process and evaluate \sysname at 5, 10,
20, and 40 req/s, reporting 10 req/s here and the others in
Appendix~\ref{sec:additional-evaluation}.
Each run uses 10 warm-up and 1,000 measured requests.
We set vLLM's concurrency limit, \texttt{max\_num\_seqs}, to 256 to match the
largest offline $B_{\mathrm{cfg}}$, or 512 to increase KV pressure.
The realized $B_{\mathrm{act}}$ is bounded by this limit, the batch supported by KV capacity, and
the number of ready requests.
At 10 req/s, all
systems saturate at limit 256; at 512, only \sysname on GLM-4.5 and Qwen3-Next
remains unsaturated.

\begin{figure}[!t]
\centering
\includegraphics[width=0.92\linewidth]{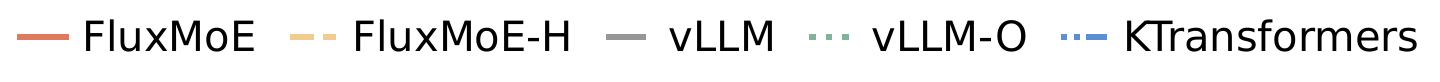}\par
\begin{tabular}{@{}c@{}c@{}c@{}}
\includegraphics[width=0.325\linewidth]{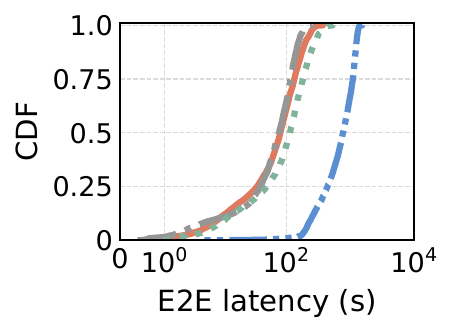} &
\includegraphics[width=0.325\linewidth]{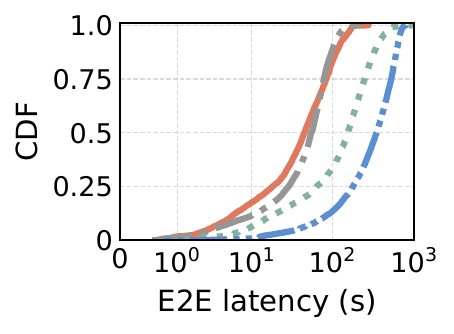} &
\includegraphics[width=0.325\linewidth]{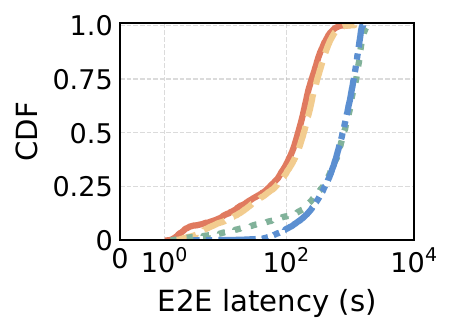} \\[-3pt]
\multicolumn{3}{c}{\small (a) Concurrency limit = 256} \\[1pt]
\includegraphics[width=0.325\linewidth]{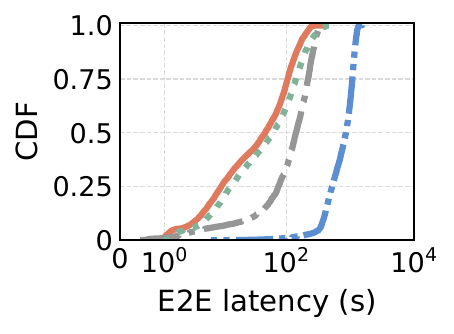} &
\includegraphics[width=0.325\linewidth]{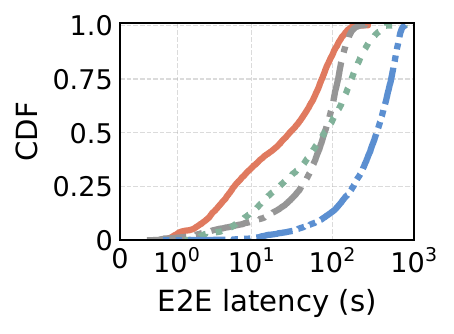} &
\includegraphics[width=0.325\linewidth]{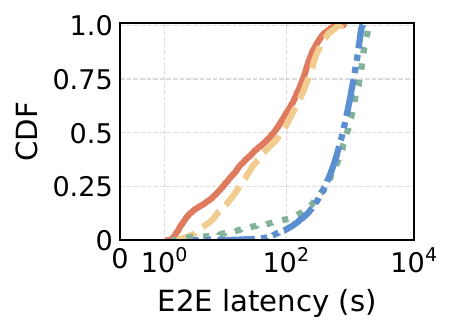} \\[-3pt]
\multicolumn{3}{c}{\small (b) Concurrency limit = 512}
\end{tabular}
\vspace{-9pt}
\captionof{figure}{(Exp\#7) Online serving performance at 10 req/s. From
left to right: GLM-4.5, Qwen3-Next, and Mixtral.}
\label{fig:exp7}
\vspace{-12pt}
\end{figure}

Figure~\ref{fig:exp7} shows the E2E-latency CDFs. Across all three models,
\sysname outperforms KTransformers because it avoids CPU expert computation
and result merging. For GLM-4.5 at a concurrency limit of 256, \sysname's P50 is 4.6\%
above vLLM's since vLLM's resident weights incur no materialization.
Raising the concurrency limit from 256 to 512 allows \sysname to exploit its
larger KV capacity and sustain a larger $\overline{B}_{\mathrm{act}}$ than vLLM.
The resulting batching benefit outweighs materialization, lowering P50 by
67.3\% versus vLLM and 24.9\% versus vLLM-O.
The latter frees KV memory but still incurs whole-layer PCIe prefetching.
\sysname avoids manually switching between vLLM's
weight-resident mode and vLLM-O's offloading mode as online conditions change.

For Mixtral, where vLLM cannot fit, \sysname lowers P50 by 18.7\% (24.1\%)
over \sysname-H at limit 256 (512).
Since both provide the same KV capacity, \sysname's gain comes
from bandwidth-balanced allocation, which
reduces materialization overhead.
Across 5, 20, and 40 req/s,
\sysname has the lowest P50 in all three models with a limit of 512,
matching the same trend observed at 10 req/s (Appendix~\ref{sec:additional-evaluation}).

\vspace{-3.5pt}

\section{Related Work}
\label{sec:related}

\para{MoE expert offloading.}
Earlier dense-model systems place parameters across GPU, CPU, and NVMe to run
models beyond GPU capacity \cite{sheng23,rajbhandari21}. In
MoE models,
loading experts from lower-tier memory creates massive PCIe traffic and makes
bandwidth a throughput bottleneck \cite{xue24,kamahori25,cao25}.
MoE systems instead offload experts through four approaches:
(i) CPU-GPU co-inference systems execute expert computation on both devices
\cite{zhong25hybri,chen25,kamahori25} or pipeline computation with weight
transfers \cite{cao25};
(ii) Prediction-based methods exploit request histories \cite{xue24,yu2026},
cross-layer routing signals \cite{fang25}, or learned predictors
\cite{zhong25} to cache and prefetch experts; some further combine prediction
with adaptive gating \cite{zhong24} or task-aware residency \cite{tairin25};
(iii) Model-level approaches modify routing to reveal future expert choices
\cite{hwang24} or fine-tune it to concentrate expert demand into a smaller
working set \cite{raje26};
and (iv) Batched-inference methods improve cache reuse and overlap expert loading
through priority-based expert caching \cite{li25diffmoe} or expert-aware
multi-batch scheduling \cite{fang25klotski}.
\sysname keeps expert computation on GPUs and preserves routing, avoiding CPU
expert-compute, output-merge, and prediction-miss overheads.
Using the fixed layer order, \sysname materializes next-layer experts from
compressed GPU memory and host DRAM concurrently, reducing PCIe stalls while
freeing GPU memory for the KV cache.

\para{KV-cache management in LLMs.}
KV-cache management follows two lines of work. The first optimizes cache layout
and reuse within GPU memory to reduce fragmentation and redundant storage
\cite{kwon23,zheng24,ye24,ye25}; the second moves KV state across devices and
storage tiers to expand capacity or support disaggregated serving
\cite{liu24,qin24,zhong24distserve}.
All assume model weights are persistent GPU-resident state;
\sysname breaks this assumption by making experts dynamic
and evictable, freeing memory for the KV cache these systems manage.

\section{Conclusion}

\sysname is an MoE serving system that decouples experts from persistent
GPU residency via expert paging.
It employs PagedTensor, a two-tier memory hierarchy,
and a budget-aware residency planner. 
Experiments on NVIDIA GPUs show that \sysname achieves up to
7.2$\times$ vLLM's throughput and 79.0\% lower TPOT for GLM-4.5, and
4.3$\times$ the throughput of KTransformers for Mixtral-8$\times$7B-Instruct,
when vLLM cannot fit, without measurable model-quality loss.

\end{sloppypar}

\bibliographystyle{ACM-Reference-Format}
\bibliography{reference}

@article{qwen3technicalreport,
  title={Qwen3 technical report}, 
  author={An Yang and Anfeng Li and Baosong Yang and Beichen Zhang and Binyuan Hui and Bo Zheng and Bowen Yu and Chang Gao and Chengen Huang and Chenxu Lv and Chujie Zheng and Dayiheng Liu and Fan Zhou and Fei Huang and Feng Hu and Hao Ge and Haoran Wei and Huan Lin and Jialong Tang and Jian Yang and Jianhong Tu and Jianwei Zhang and Jianxin Yang and Jiaxi Yang and Jing Zhou and Jingren Zhou and Junyang Lin and Kai Dang and Keqin Bao and Kexin Yang and Le Yu and Lianghao Deng and Mei Li and Mingfeng Xue and Mingze Li and Pei Zhang and Peng Wang and Qin Zhu and Rui Men and Ruize Gao and Shixuan Liu and Shuang Luo and Tianhao Li and Tianyi Tang and Wenbiao Yin and Xingzhang Ren and Xinyu Wang and Xinyu Zhang and Xuancheng Ren and Yang Fan and Yang Su and Yichang Zhang and Yinger Zhang and Yu Wan and Yuqiong Liu and Zekun Wang and Zeyu Cui and Zhenru Zhang and Zhipeng Zhou and Zihan Qiu},
  year={2025},
  journal={arXiv preprint arXiv:2505.09388},
  url={https://arxiv.org/abs/2505.09388},
}

@article{qin24,
  title={Mooncake: A {KVCache-centric} disaggregated architecture for {LLM} serving},
  author={Qin, Ruoyu and Li, Zheming and He, Weiran and Cui, Jialei and Tang, Heyi and Ren, Feng and Ma, Teng and Cai, Shangming and Zhang, Yineng and Zhang, Mingxing and Wu, Yongwei and Zheng, Weimin and Xu, Xinran},
  journal={ACM Trans. on Storage},
  year={2026},
  doi={10.1145/3773772}
}

@inproceedings{zhong24distserve,
  title={{DistServe}: Disaggregating prefill and decoding for goodput-optimized large language model serving},
  author={Zhong, Yinmin and Liu, Shengyu and Chen, Junda and Hu, Jianbo and Zhu, Yibo and Liu, Xuanzhe and Jin, Xin and Zhang, Hao},
  booktitle={Proc. of USENIX OSDI},
  year={2024},
  url={https://www.usenix.org/conference/osdi24/presentation/zhong-yinmin}
}

@article{Bai2025KimiKO,
  title={{Kimi K2}: Open agentic intelligence},
  author={Kimi Team and Yifan Bai and Yiping Bao and Guanduo Chen and Jiahao Chen and Ningxin Chen and Ruijue Chen and Yanru Chen and Yuankun Chen and Yutian Chen and Zhuofu Chen and Jialei Cui and Haochen Ding and Meng-xiao Dong and Angang Du and Chenzhuang Du and Dikang Du and Yulun Du and Yu Fan and Yichen Feng and Kelin Fu and Bofei Gao and Hongcheng Gao and Peizhong Gao and Tong Gao and Xinran Gu and Longyu Guan and Haiqing Guo and Jia-Xing Guo and Hao-Xing Hu and Xiaoru Hao and Tian He and Weiran He and Wen He and Chao Hong and Yan-Ni Hu and Zhenxing Hu and Weixiao Huang and Zhiqi Huang and Zihao Huang and Tao Jiang and Zhejun Jiang and Xinyi Jin and Yongsheng Kang and Guokun Lai and Cheng Li and Fang Li and Haoyang Li and Ming Li and Wentao Li and Yanhao Li and Yiwei Li and Zhaowei Li and Zheming Li and Hong-Li Lin and Xiaohan Lin and Zongyu Lin and Chengyi Liu and Chenyu Liu and Hongzhang Liu and Jingyuan Liu and Junqi Liu and Liang Liu and Shaowei Liu and T. Y. Liu and Tian-Bo Liu and Weizhou Liu and Yangyang Liu and Yibo Liu and Yiping Liu and Yue Liu and Zhengying Liu and Enzhe Lu and Li Lu and Shen Ma and Xinyu Ma and Yi-Xuan Ma and Shaoguang Mao and Jie Mei and Xin Men and Yibo Miao and Siyuan Pan and Yebo Peng and Ruoyu Qin and Bowen Qu and Zeyu Shang and Li-Na Shi and Sheng-Rong Shi and Feifan Song and Jian-Fei Su and Zhen-Xin Su and Xinjie Sun and Flood Sung and Heyi Tang and Ji-Hua Tao and Qi Teng and Chensi Wang and Dinglu Wang and Feng Wang and Haiming Wang and Jianzhou Wang and Jiaxing Wang and Jinhong Wang and Shengjie Wang and Shuyi Wang and Yao Wang and Yejie Wang and Yiqin Wang and Yuxin Wang and Yuzhi Wang and Zhaoji Wang and Zhengtao Wang and Zhexu Wang and Chu Wei and Qi-Feng Wei and Wenhao Wu and Xingzhe Wu and Yuxin Wu and Chenjun Xiao and Xiao-Ming Xie and Weiming Xiong and Boyu Xu and Jing Xu and Jinjing Xu and L. H. Xu and Lin Xu and Suting Xu and Weixin Xu and Xinran Xu and Yangchuan Xu and Zi-Yang Xu and Junjie Yan and Yuzi Yan and Xiaofei Yang and Ying Yang and Zhengqi Yang and Zhilin Yang and Zonghan Yang and Haotian Yao and Xingcheng Yao and Wen-guang Ye and Zhuorui Ye and Bohong Yin and Long Yu and Enming Yuan and Hongbang Yuan and Mengjie Yuan and Haobing Zhan and Dehao Zhang and Hao Zhang and Wanlu Zhang and Xiaobin Zhang and Yangkun Zhang and Yizhi Zhang and Yongting Zhang and Yu Zhang and Yutao Zhang and Yutong Zhang and Zheng Zhang and Hao-Dong Zhao and Yikai Zhao and Huabin Zheng and Shao Jian Zheng and Jianren Zhou and Xinyu Zhou and Zaida Zhou and Zhengxin Zhu and Weiyu Zhuang and Xinxing Zu},
  year={2025},
  journal={arXiv preprint arXiv:2507.20534},
  url={https://arxiv.org/abs/2507.20534},
}

@article{DeepSeekAI2025DeepSeekR1IR,
  title={{DeepSeek-R1} incentivizes reasoning in {LLMs} through reinforcement learning},
  author={DeepSeek-AI and Daya Guo and Dejian Yang and Haowei Zhang and Jun-Mei Song and Ruoyu Zhang and Runxin Xu and Qihao Zhu and Shirong Ma and Peiyi Wang and Xiaoling Bi and Xiaokang Zhang and Xingkai Yu and Yu Wu and Z. F. Wu and Zhibin Gou and Zhihong Shao and Zhuoshu Li and Ziyi Gao and Aixin Liu and Bing Xue and Bing-Li Wang and Bochao Wu and Bei Feng and Chengda Lu and Chenggang Zhao and Chengqi Deng and Chenyu Zhang and Chong Ruan and Damai Dai and Deli Chen and Dong-Li Ji and Erhang Li and Fangyun Lin and Fucong Dai and Fuli Luo and Guangbo Hao and Guanting Chen and Guowei Li and H. Zhang and Han Bao and Hanwei Xu and Haocheng Wang and Honghui Ding and Huajian Xin and Huazuo Gao and Hui Qu and Hui Li and Jianzhong Guo and Jiashi Li and Jiawei Wang and JingChang Chen and Jingyang Yuan and Junjie Qiu and Junlong Li and Jiong Cai and Jiaqi Ni and Jian Liang and Jin Chen and Kai Dong and Kai Hu and Kaige Gao and Kang Guan and Kexin Huang and Kuai Yu and Lean Wang and Lecong Zhang and Liang Zhao and Litong Wang and Liyue Zhang and Lei Xu and Leyi Xia and Mingchuan Zhang and Minghua Zhang and M. Tang and Meng Li and Miaojun Wang and Mingming Li and Ning Tian and Panpan Huang and Peng Zhang and Qiancheng Wang and Qinyu Chen and Qiushi Du and Ruiqi Ge and Ruisong Zhang and Ruizhe Pan and Runji Wang and R. J. Chen and Ruiqi Jin and Ruyi Chen and Shanghao Lu and Shangyan Zhou and Shanhuang Chen and Shengfeng Ye and Shiyu Wang and Shuiping Yu and Shunfeng Zhou and Shuting Pan and S. S. Li and Shuang Zhou and Shao-Kang Wu and Tao Yun and Tian Pei and Tianyu Sun and T. Wang and Wangding Zeng and Wanjia Zhao and Wen Liu and Wenfeng Liang and Wenjun Gao and Wen-Xia Yu and Wentao Zhang and Wangding Xiao and Wei An and Xiaodong Liu and Xiaohan Wang and Xiaokang Chen and Xiaotao Nie and Xin Cheng and Xin Liu and Xin Xie and Xingchao Liu and Xinyu Yang and Xinyuan Li and Xuecheng Su and Xuheng Lin and X. Q. Li and Xiangyu Jin and Xi-Cheng Shen and Xiaosha Chen and Xiaowen Sun and Xiaoxiang Wang and Xinnan Song and Xinyi Zhou and Xianzu Wang and Xinxia Shan and Y. K. Li and Y. Q. Wang and Y. X. Wei and Yang Zhang and Yanhong Xu and Yao Li and Yao Zhao and Yaofeng Sun and Yaohui Wang and Yi Yu and Yichao Zhang and Yifan Shi and Yi Xiong and Ying He and Yishi Piao and Yisong Wang and Yixuan Tan and Yiyang Ma and Yiyuan Liu and Yongqiang Guo and Yuan Ou and Yuduan Wang and Yue Gong and Yu-Jing Zou and Yujia He and Yunfan Xiong and Yu-Wei Luo and Yu-mei You and Yuxuan Liu and Yuyang Zhou and Y. X. Zhu and Yanping Huang and Yao Li and Yi Zheng and Yuchen Zhu and Yunxiang Ma and Ying Tang and Yukun Zha and Yuting Yan and Zehui Ren and Zehui Ren and Zhangli Sha and Zhe Fu and Zhean Xu and Zhenda Xie and Zhen-guo Zhang and Zhewen Hao and Zhicheng Ma and Zhigang Yan and Zhiyu Wu and Zihui Gu and Zijia Zhu and Zijun Liu and Zi-An Li and Ziwei Xie and Ziyang Song and Zizheng Pan and Zhen Huang and Zhipeng Xu and Zhongyu Zhang and Zhen Zhang},
  journal={Nature},
  year={2025},
  volume={645},
  pages={633 - 638},
  doi={10.1038/s41586-025-09422-z}
}

@inproceedings{merity17,
  title={Pointer sentinel mixture models},
  author={Merity, Stephen and Xiong, Caiming and Bradbury, James and Socher, Richard},
  booktitle={International Conference on Learning Representations},
  year={2017},
  url={https://openreview.net/forum?id=Byj72udxe}
}

@inproceedings{sheng23,
  title={{FlexGen}: High-throughput generative inference of large language models with a single {GPU}},
  author={Sheng, Ying and Zheng, Lianmin and Yuan, Binhang and Li, Zhuohan and Ryabinin, Max and Chen, Beidi and Liang, Percy and R{\'e}, Christopher and Stoica, Ion and Zhang, Ce},
  booktitle={International Conference on Machine Learning},
  year={2023},
  url={https://proceedings.mlr.press/v202/sheng23a.html},
}

@article{tong24,
  title={{DART-Math}: Difficulty-aware rejection tuning for mathematical problem-solving},
  author={Tong, Yuxuan and Zhang, Xiwen and Wang, Rui and Wu, Ruidong and He, Junxian},
  journal={Advances in Neural Information Processing Systems},
  volume={37},
  pages={7821--7846},
  year={2024},
  doi={10.52202/079017-0251},
}

@inproceedings{sheng25,
  title={{HybridFlow}: A flexible and efficient {RLHF} framework},
  author={Sheng, Guangming and Zhang, Chi and Ye, Zilingfeng and Wu, Xibin and Zhang, Wang and Zhang, Ru and Peng, Yanghua and Lin, Haibin and Wu, Chuan},
  booktitle={Proc. of ACM EuroSys},
  year={2025},
  doi={10.1145/3689031.3696075},
}

@inproceedings{aggarwal25,
  title={{L1}: Controlling how long a reasoning model thinks with reinforcement learning},
  author={Aggarwal, Pranjal and Welleck, Sean},
  booktitle={Second Conference on Language Modeling},
  year={2025},
  url={https://openreview.net/forum?id=4jdIxXBNve},
}

@inproceedings{kwon23,
  title={Efficient memory management for large language model serving with {PagedAttention}},
  author={Kwon, Woosuk and Li, Zhuohan and Zhuang, Siyuan and Sheng, Ying and Zheng, Lianmin and Yu, Cody Hao and Gonzalez, Joseph and Zhang, Hao and Stoica, Ion},
  booktitle={Proc. of ACM SOSP},
  year={2023},
  doi={10.1145/3600006.3613165}
}

@article{zheng24,
  title={{SGLang}: Efficient execution of structured language model programs},
  author={Zheng, Lianmin and Yin, Liangsheng and Xie, Zhiqiang and Sun, Chuyue Livia and Huang, Jeff and Yu, Cody Hao and Cao, Shiyi and Kozyrakis, Christos and Stoica, Ion and Gonzalez, Joseph E and Barrett, Clark and Sheng, Ying},
  journal={Advances in neural information processing systems},
  volume={37},
  pages={62557--62583},
  year={2024},
  doi={10.52202/079017-2000}
}

@inproceedings{yu22,
  title={Orca: A distributed serving system for {Transformer-Based} generative models},
  author={Yu, Gyeong-In and Jeong, Joo Seong and Kim, Geon-Woo and Kim, Soojeong and Chun, Byung-Gon},
  booktitle={Proc. of USENIX OSDI},
  year={2022},
  url={https://www.usenix.org/conference/osdi22/presentation/yu}
}

@article{ye24,
  title={{ChunkAttention}: Efficient self-attention with prefix-aware {KV} cache and two-phase partition},
  author={Ye, Lu and Tao, Ze and Huang, Yong and Li, Yang},
  journal={Annual Meeting of the Association for Computational Linguistics},
  year={2024},
  doi={10.18653/v1/2024.acl-long.623}
}

@inproceedings{liu24,
  title={{CacheGen}: {KV} cache compression and streaming for fast large language model serving},
  author={Liu, Yuhan and Li, Hanchen and Cheng, Yihua and Ray, Siddhant and Huang, Yuyang and Zhang, Qizheng and Du, Kuntai and Yao, Jiayi and Lu, Shan and Ananthanarayanan, Ganesh and Maire, Michael and Hoffmann, Henry and Holtzman, Ari and Jiang, Junchen},
  booktitle={Proc. of ACM SIGCOMM},
  year={2024},
  doi={10.1145/3651890.3672274}
}

@inproceedings{hershcovitch25,
  title={{ZipNN}: Lossless compression for {AI} models},
  author={Hershcovitch, Moshik and Wood, Andrew and Choshen, Leshem and Girmonsky, Guy and Leibovitz, Roy and Ozeri, Or and Ennmouri, Ilias and Malka, Michal and Chin, Peter and Sundararaman, Swaminathan and Harnik, Danny},
  booktitle={Proc. of IEEE CLOUD},
  year={2025},
  doi={10.1109/CLOUD67622.2025.00028}
}

@article{heilper25,
  title={Lossless compression of neural network components: Weights, checkpoints, and {K/V} caches in low-precision formats},
  author={Heilper, Anat and Singer, Doron},
  journal={arXiv preprint arXiv:2508.19263},
  year={2025},
  url={https://arxiv.org/abs/2508.19263},
}

@inproceedings{zhang25,
  title={70\% size, 100\% accuracy: Lossless {LLM} compression for efficient {GPU} inference via dynamic-length float},
  author={Zhang, Tianyi and Hariri, Mohsen and Zhong, Shaochen and Chaudhary, Vipin and Sui, Yang and Hu, Xia and Shrivastava, Anshumali},
  booktitle={Annual Conference on Neural Information Processing Systems},
  year={2025},
  doi={10.52202/085713-3306}
}

@inproceedings{fan26,
  title={{ZipServ}: Fast and memory-efficient {LLM} inference with hardware-aware lossless compression},
  author={Fan, Ruibo and Yu, Xiangrui and Pan, Xinglin and Li, Zeyu and Luo, Weile and Wang, Qiang and Wang, Wei and Chu, Xiaowen},
  booktitle={Proc. of ACM ASPLOS},
  year={2026},
  doi={10.1145/3779212.3790250}
}

@article{xue24,
  title={{Moe-Infinity}: Efficient {MoE} inference on personal machines with sparsity-aware expert cache},
  author={Xue, Leyang and Fu, Yao and Lu, Zhan and Mai, Luo and Marina, Mahesh},
  journal={arXiv preprint arXiv:2401.14361},
  year={2024},
  url={https://arxiv.org/abs/2401.14361},
}

@misc{shareGPT,
  title = {{ShareGPT} datasets},
  howpublished = {\url{https://huggingface.co/collections/bunnycore/sharegpt-datasets-66fa831dcee14c587f1e6d1c}},
}

@misc{cudavmm,
  title = {Virtual memory management {APIs} in {CUDA} programming guide},
  howpublished = {\url{https://docs.nvidia.com/cuda/cuda-programming-guide/04-special-topics/virtual-memory-management.html}},
}

@article{glm45,
  title={{GLM-4.5}: Agentic, reasoning, and coding {(ARC)} foundation models}, 
  author={GLM-4.5 Team and Aohan Zeng and Xin Lv and Qinkai Zheng and Zhenyu Hou and Bin Chen and Chengxing Xie and Cunxiang Wang and Da Yin and Hao Zeng and Jiajie Zhang and Kedong Wang and Lucen Zhong and Mingdao Liu and Rui Lu and Shulin Cao and Xiaohan Zhang and Xuancheng Huang and Yao Wei and Yean Cheng and Yifan An and Yilin Niu and Yuanhao Wen and Yushi Bai and Zhengxiao Du and Zihan Wang and Zilin Zhu and Bohan Zhang and Bosi Wen and Bowen Wu and Bowen Xu and Can Huang and Casey Zhao and Changpeng Cai and Chao Yu and Chen Li and Chendi Ge and Chenghua Huang and Chenhui Zhang and Chenxi Xu and Chenzheng Zhu and Chuang Li and Congfeng Yin and Daoyan Lin and Dayong Yang and Dazhi Jiang and Ding Ai and Erle Zhu and Fei Wang and Gengzheng Pan and Guo Wang and Hailong Sun and Haitao Li and Haiyang Li and Haiyi Hu and Hanyu Zhang and Hao Peng and Hao Tai and Haoke Zhang and Haoran Wang and Haoyu Yang and He Liu and He Zhao and Hongwei Liu and Hongxi Yan and Huan Liu and Huilong Chen and Ji Li and Jiajing Zhao and Jiamin Ren and Jian Jiao and Jiani Zhao and Jianyang Yan and Jiaqi Wang and Jiayi Gui and Jiayue Zhao and Jie Liu and Jijie Li and Jing Li and Jing Lu and Jingsen Wang and Jingwei Yuan and Jingxuan Li and Jingzhao Du and Jinhua Du and Jinxin Liu and Junkai Zhi and Junli Gao and Ke Wang and Lekang Yang and Liang Xu and Lin Fan and Lindong Wu and Lintao Ding and Lu Wang and Man Zhang and Minghao Li and Minghuan Xu and Mingming Zhao and Mingshu Zhai and Pengfan Du and Qian Dong and Shangde Lei and Shangqing Tu and Shangtong Yang and Shaoyou Lu and Shijie Li and Shuang Li and Shuang-Li and Shuxun Yang and Sibo Yi and Tianshu Yu and Wei Tian and Weihan Wang and Wenbo Yu and Weng Lam Tam and Wenjie Liang and Wentao Liu and Xiao Wang and Xiaohan Jia and Xiaotao Gu and Xiaoying Ling and Xin Wang and Xing Fan and Xingru Pan and Xinyuan Zhang and Xinze Zhang and Xiuqing Fu and Xunkai Zhang and Yabo Xu and Yandong Wu and Yida Lu and Yidong Wang and Yilin Zhou and Yiming Pan and Ying Zhang and Yingli Wang and Yingru Li and Yinpei Su and Yipeng Geng and Yitong Zhu and Yongkun Yang and Yuhang Li and Yuhao Wu and Yujiang Li and Yunan Liu and Yunqing Wang and Yuntao Li and Yuxuan Zhang and Zezhen Liu and Zhen Yang and Zhengda Zhou and Zhongpei Qiao and Zhuoer Feng and Zhuorui Liu and Zichen Zhang and Zihan Wang and Zijun Yao and Zikang Wang and Ziqiang Liu and Ziwei Chai and Zixuan Li and Zuodong Zhao and Wenguang Chen and Jidong Zhai and Bin Xu and Minlie Huang and Hongning Wang and Juanzi Li and Yuxiao Dong and Jie Tang},
  journal={arXiv preprint arXiv:2508.06471},
  year={2025},
  url={https://arxiv.org/abs/2508.06471},
}

@article{lin24,
  title={{AWQ}: Activation-aware weight quantization for on-device {LLM} compression and acceleration},
  author={Lin, Ji and Tang, Jiaming and Tang, Haotian and Yang, Shang and Chen, Wei-Ming and Wang, Wei-Chen and Xiao, Guangxuan and Dang, Xingyu and Gan, Chuang and Han, Song},
  journal={Proceedings of Machine Learning and Systems},
  volume={6},
  pages={87--100},
  year={2024},
  url={https://proceedings.mlsys.org/paper_files/paper/2024/hash/42a452cbafa9dd64e9ba4aa95cc1ef21-Abstract-Conference.html}
}

@article{frantar23,
  title={{GPTQ}: Accurate post-training quantization for generative pre-trained transformers},
  author={Frantar, Elias and Ashkboos, Saleh and Hoefler, Torsten and Alistarh, Dan},
  journal={International Conference on Learning Representations},
  year={2023},
  url={https://arxiv.org/abs/2210.17323}
}

@article{huffman07,
  title={A method for the construction of minimum-redundancy codes},
  author={Huffman, David A},
  journal={Proceedings of the IRE},
  volume={40},
  number={9},
  pages={1098--1101},
  year={1952},
  doi={10.1109/JRPROC.1952.273898}
}

@inproceedings{zhong25hybri,
  title={{HybriMoE}: Hybrid {CPU-GPU} Scheduling and cache management for efficient {MoE} inference},
  author={Zhong, Shuzhang and Sun, Yanfan and Liang, Ling and Wang, Runsheng and Huang, Ru and Li, Meng},
  booktitle={Proc. of ACM/IEEE DAC},
  year={2025},
  doi={10.1109/DAC63849.2025.11133274}
}

@inproceedings{cao25,
  title={{MoE-Lightning}: High-throughput {MoE} inference on memory-constrained {GPUs}},
  author={Cao, Shiyi and Liu, Shu and Griggs, Tyler and Schafhalter, Peter and Liu, Xiaoxuan and Sheng, Ying and Gonzalez, Joseph E and Zaharia, Matei and Stoica, Ion},
  booktitle={Proc. of ACM ASPLOS},
  year={2025},
  doi={10.1145/3669940.3707267}
}

@inproceedings{kamahori25,
  title={{Fiddler}: {CPU-GPU} orchestration for fast inference of mixture-of-experts models},
  author={Kamahori, Keisuke and Tang, Tian and Gu, Yile and Zhu, Kan and Kasikci, Baris},
  booktitle={International Conference on Learning Representations},
  year={2025},
  url={https://openreview.net/forum?id=N5fVv6PZGz}
}

@inproceedings{fang25klotski,
  title={{Klotski}: Efficient mixture-of-expert inference via expert-aware multi-batch pipeline},
  author={Fang, Zhiyuan and Huang, Yuegui and Hong, Zicong and Lyu, Yufeng and Chen, Wuhui and Yu, Yue and Yu, Fan and Zheng, Zibin},
  booktitle={Proc. of ACM ASPLOS},
  year={2025},
  doi={10.1145/3676641.3716261},
}

@inproceedings{zhong24,
  title={{AdapMoE}: Adaptive sensitivity-based expert gating and management for efficient {MoE} inference},
  author={Zhong, Shuzhang and Liang, Ling and Wang, Yuan and Wang, Runsheng and Huang, Ru and Li, Meng},
  booktitle={Proc. of IEEE/ACM ICCAD},
  year={2024},
  doi={10.1145/3676536.3676741}
}

@inproceedings{hwang24,
  title={Pre-gated {MoE}: An algorithm-system co-design for fast and scalable mixture-of-expert inference},
  author={Hwang, Ranggi and Wei, Jianyu and Cao, Shijie and Hwang, Changho and Tang, Xiaohu and Cao, Ting and Yang, Mao},
  booktitle={Proc. of ACM/IEEE ISCA},
  year={2024},
  doi={10.1109/ISCA59077.2024.00078}
}

@inproceedings{fang25,
  title={{Fate}: Fast edge inference of mixture-of-experts models via cross-layer gate},
  author={Fang, Zhiyuan and Yu, Xingfan and Huang, Yuegui and Hong, Zicong and Lyu, Yufeng and Chen, Wuhui and Yu, Yue and Yu, Fan},
  booktitle={Proc. of ACM WWW},
  year={2026},
  doi={10.1145/3774904.3792527}
}

@article{tairin25,
  title={{eMoE}: Task-aware memory efficient mixture-of-experts-based ({MoE}) model inference},
  author={Tairin, Suraiya and Mahmud, Shohaib and Shen, Haiying and Iyer, Anand},
  journal={arXiv preprint arXiv:2503.06823},
  year={2025},
  url={https://arxiv.org/abs/2503.06823},
}

@inproceedings{li25diffmoe,
  title={{Diff-MoE}: Efficient batched {MoE} inference with priority-driven differential expert caching},
  author={Li, Kexin and Huang, Wenkan and Wang, Qinggang and Zheng, Long and Liao, Xiaofei and Jin, Hai and Xue, Jingling},
  booktitle={Proc. of ACM SC},
  year={2025},
  doi={10.1145/3712285.3759903}
}

@inproceedings{yu2026,
  title={Taming latency-memory trade-off in {MoE}-based {LLM} serving via fine-grained expert offloading},
  author={Yu, Hanfei and Cui, Xingqi and Zhang, Hong and Wang, Hao and Wang, Hao},
  booktitle={Proc. of ACM EuroSys},
  year={2026},
  doi={10.1145/3767295.3769319}
}

@inproceedings{chen25,
  title={{KTransformers}: Unleashing the full potential of {CPU/GPU} hybrid inference for MoE models},
  author={Chen, Hongtao and Xie, Weiyu and Zhang, Boxin and Tang, Jingqi and Wang, Jiahao and Dong, Jianwei and Chen, Shaoyuan and Yuan, Ziwei and Lin, Chen and Qiu, Chengyu and Zhu, Yuening and Ou, Qingliang and Liao, Jiaqi and Chen, Xianglin and Ai, Zhiyuan and Wu, Yongwei and Zhang, Mingxing},
  booktitle={Proc. of ACM SOSP},
  year={2025},
  doi={10.1145/3731569.3764843}
}

@misc{github,
  title={{KTransformers} GitHub repository},
  howpublished={\url{https://github.com/kvcache-ai/ktransformers}},
}

@misc{config,
  title={{KT-Kernel} configuration guide for {KTransformers}},
  howpublished={\url{https://github.com/kvcache-ai/ktransformers/blob/main/kt-kernel/README.md\#kt-kernel-parameters}},
}

@inproceedings{zhong25,
  title={{ExpertFlow}: Efficient mixture-of-experts inference via predictive expert caching and token scheduling},
  author={He, Xin and Zhang, Shunkang and Tang, Kaijie and Shi, Shaohuai and Wang, Yuxin and Zeng, Zihao and Tang, Zhenheng and Chu, Xiaowen and Yin, Haiyan and Tsang, Ivor W. and Ong, Yew Soon},
  booktitle={Proc. of ACM/IEEE DAC},
  year={2026},
  url={https://arxiv.org/abs/2410.17954}
}

@article{yang2026,
  title={{ZipMoE}: Efficient on-device {MoE} serving via lossless compression and cache-affinity scheduling},
  author={Yang, Yuchen and Zhao, Yaru and Yang, Pu and Wang, Shaowei and Zhou, Zhi-Hua},
  journal={arXiv preprint arXiv:2601.21198},
  year={2026},
  url={https://arxiv.org/abs/2601.21198},
}

@article{jiang24,
  title={Mixtral of experts}, 
  author={Albert Q. Jiang and Alexandre Sablayrolles and Antoine Roux and Arthur Mensch and Blanche Savary and Chris Bamford and Devendra Singh Chaplot and Diego de las Casas and Emma Bou Hanna and Florian Bressand and Gianna Lengyel and Guillaume Bour and Guillaume Lample and Lélio Renard Lavaud and Lucile Saulnier and Marie-Anne Lachaux and Pierre Stock and Sandeep Subramanian and Sophia Yang and Szymon Antoniak and Teven Le Scao and Théophile Gervet and Thibaut Lavril and Thomas Wang and Timothée Lacroix and William El Sayed},
  year={2024},
  journal={arXiv preprint arXiv:2401.04088},
  url={https://arxiv.org/abs/2401.04088},
}

@inproceedings{rajbhandari21,
  title={{ZeRO-Infinity}: Breaking the {GPU} memory wall for extreme scale deep learning},
  author={Rajbhandari, Samyam and Ruwase, Olatunji and Rasley, Jeff and Smith, Shaden and He, Yuxiong},
  booktitle={Proc. of ACM SC},
  year={2021},
  doi={10.1145/3458817.3476205}
}

@article{raje26,
  title={{MELINOE}: Fine-tuning enables memory-efficient inference for mixture-of-experts models},
  author={Raje, Arian and Nayak, Anupam and Joshi, Gauri},
  journal={arXiv preprint arXiv:2602.11192},
  year={2026},
  url={https://arxiv.org/abs/2602.11192},
}

@inproceedings{xiang25,
  author = {Xiang, Yuxing and Li, Xue and Qian, Kun and Yang, Yufan and Zhu, Diwen and Yu, Wenyuan and Zhai, Ennan and Liu, Xuanzhe and Jin, Xin and Zhou, Jingren},
  title = {Aegaeon: Effective {GPU} pooling for concurrent {LLM} Serving on the market},
  year = {2025},
  booktitle = {Proc. of ACM SOSP},
  doi = {10.1145/3731569.3764815},
}

@inproceedings{jeong25,
  author = {Jeong, Jinwoo and Ahn, Jeongseob},
  title = {Accelerating {LLM} serving for multi-turn dialogues with efficient resource management},
  year = {2025},
  booktitle = {Proc. of ACM ASPLOS},
  doi = {10.1145/3676641.3716245},
}

@inproceedings{ye25,
  author = {Ye, Zihao and Chen, Lequn and Lai, Ruihang and Lin, Wuwei and Zhang, Yineng and Wang, Stephanie and Chen, Tianqi and Kasikci, Baris and Grover, Vinod and Krishnamurthy, Arvind and Ceze, Luis},
  booktitle = {Proceedings of Machine Learning and Systems},
  title = {{FlashInfer}: Efficient and customizable attention engine for {LLM} inference serving},
  year = {2025},
  url = {https://proceedings.mlsys.org/paper_files/paper/2025/hash/dbf02b21d77409a2db30e56866a8ab3a-Abstract-Conference.html}
}

@inproceedings{zhu25,
  title={{NanoFlow}: Towards optimal large language model serving throughput},
  author={Kan Zhu and Yufei Gao and Yilong Zhao and Liangyu Zhao and Gefei Zuo and Yile Gu and Dedong Xie and Tian Tang and Qinyu Xu and Zihao Ye and Keisuke Kamahori and Chien-Yu Lin and Ziren Wang and Stephanie Wang and Arvind Krishnamurthy and Baris Kasikci},
  booktitle={Proc. of USENIX OSDI},
  year={2025},
  url={https://www.usenix.org/conference/osdi25/presentation/zhu-kan}
}

@inproceedings{gao25,
  title={{Weaver}: Efficient multi-{LLM} serving with attention offloading},
  author={Shiwei Gao and Qing Wang and Shaoxun Zeng and Youyou Lu and Jiwu Shu},
  booktitle={Proc. of USENIX ATC},
  year={2025},
  url={https://www.usenix.org/conference/atc25/presentation/gao}
}

@inproceedings{tillet19,
  author = {Tillet, Philippe and Kung, H. T. and Cox, David},
  title = {Triton: An intermediate language and compiler for tiled neural network computations},
  year = {2019},
  booktitle = {Proc. of ACM MAPL},
  doi = {10.1145/3315508.3329973},
}

@article{paszke19,
  title={{PyTorch}: An imperative style, high-performance deep learning library},
  author={Adam Paszke and Sam Gross and Francisco Massa and Adam Lerer and James Bradbury and Gregory Chanan and Trevor Killeen and Zeming Lin and Natalia Gimelshein and Luca Antiga and Alban Desmaison and Andreas Köpf and Edward Yang and Zach DeVito and Martin Raison and Alykhan Tejani and Sasank Chilamkurthy and Benoit Steiner and Lu Fang and Junjie Bai and Soumith Chintala},
  journal={Advances in neural information processing systems},
  volume={32},
  pages={8026 - 8037},
  year={2019},
  url={https://proceedings.neurips.cc/paper/9015-pytorch-an-imperative-style-high-performance-deep-learning-library}
}

\clearpage
\appendix
\label{sec:appendix}

This appendix provides analyses and evaluation details omitted from the paper.
We first characterize large-batch expert activation and the bit-level
compressibility of MoE expert weights
(Appendix~\ref{sec:additional-analysis}), then supplement our
experiments with additional results
(Appendix~\ref{sec:additional-evaluation}), and finally verify that \sysname
preserves model quality
(Appendix~\ref{sec:additional-exp}).

\section{Additional Analysis}
\label{sec:additional-analysis}

\begin{figure}[t]
\centering
\begin{tabular}{@{}c@{\hspace{6pt}}c@{}}
\includegraphics[width=0.48\columnwidth]{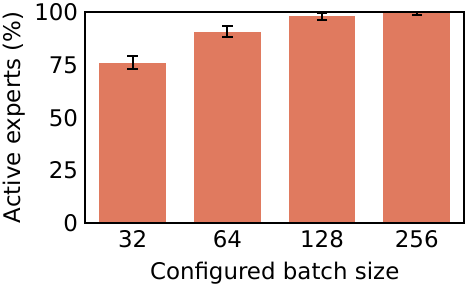} &
\includegraphics[width=0.48\columnwidth]{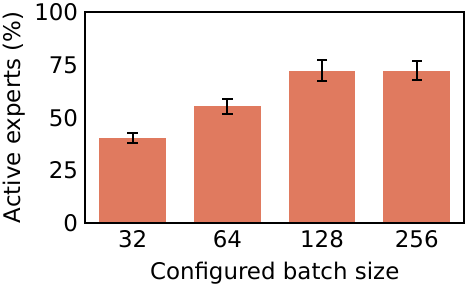}
\\[-1pt]
\multicolumn{2}{c}{\small (a) Average fractions across layers and iterations}
\end{tabular}
\begin{tabular}{@{}c@{\hspace{8pt}}c@{}}
\includegraphics[width=0.485\columnwidth]{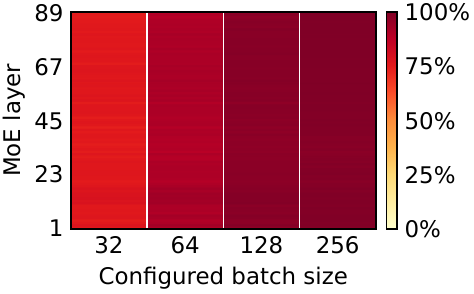} &
\includegraphics[width=0.485\columnwidth]{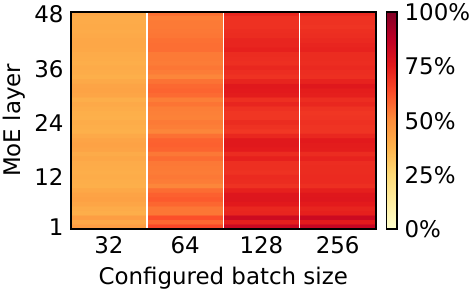}
\\[-1pt]
\multicolumn{2}{c}{\small (b) Per-layer average fractions across iterations}
\end{tabular}
\vspace{-6pt}
\captionof{figure}{Active-expert fractions during decoding for
GLM-4.5 \cite{glm45} (left) and Qwen3-Next-80B-A3B-Instruct \cite{qwen3technicalreport} (right).}
\label{fig:expert_activation}
\vspace{-6pt}
\end{figure}

\para{Large-batch expert activation.}
To quantify expert activation at large batch sizes, we run vLLM \cite{kwon23} with the model
and prompt configurations in \S\ref{subsec:setup} and profile the fraction of
routed experts activated at configured batch sizes of 32, 64, 128, and 256. We select
GLM-4.5 \cite{glm45} and Qwen3-Next-80B-A3B-Instruct \cite{qwen3technicalreport}
since they combine large routed-expert pools (160 and 512 experts per layer,
respectively) with sparse per-token routing (top-8 and top-10, respectively).
For every decoding iteration and MoE layer, we record the number of distinct routed experts
selected by the scheduled tokens. Shared experts are excluded from both the
counts and pool sizes. Since KV capacity may prevent the system from processing
the full configured batch, we measure activation only in decoding iterations whose
scheduled sequence count reaches at least 90\% of the run's maximum.
Each retained layer--iteration pair forms one sample, whose active-expert
fraction is the distinct routed-expert count divided by that layer's
routed-expert count.

Figure~\ref{fig:expert_activation}(a) pools these samples' active-expert fractions
across layers and iterations, reporting their average and standard deviation.
The average active-expert fraction rises with configured batch size, reaching
nearly 100\% for GLM-4.5
\cite{glm45} and over 70\% for Qwen3-Next-80B-A3B-Instruct
\cite{qwen3technicalreport} at a configured batch size of 256 since larger scheduled batches generate
more routing decisions. GLM-4.5 has a higher fraction since
each GLM-4.5 token selects 5\% of its expert pool, whereas each
Qwen3-Next-80B-A3B-Instruct token selects about 2\%. For Qwen3-Next-80B-A3B-Instruct, the fraction
plateaus as the configured batch size increases from 128 to 256
since limited KV capacity prevents more sequences from being scheduled.
Figure~\ref{fig:expert_activation}(b) instead averages the samples' active-expert fractions across
retained iterations separately for each layer. The active-expert fraction
increases in nearly all layers rather than only a few outliers since a larger scheduled batch
supplies more routing decisions to every layer.
These results motivate layer-wise expert streaming: large batches activate
most experts within each layer, shifting the memory-saving opportunity from
expert-level caching to cross-layer memory reuse.

\para{Expert-weight compressibility.}
To characterize the compressibility of MoE expert weights, we profile
the BFloat16 routed-expert weights of Mixtral-8$\times$7B-Instruct
\cite{jiang24} and Qwen3-Next-80B-A3B-Instruct
\cite{qwen3technicalreport} across their MoE layers. We separate
the weights into exponent and mantissa fields and compute their empirical
distributions across experts. 

\begin{figure}[t]
\centering
\begin{tabular}{@{}c@{\hspace{1pt}}c@{}}
\includegraphics[width=0.485\linewidth]{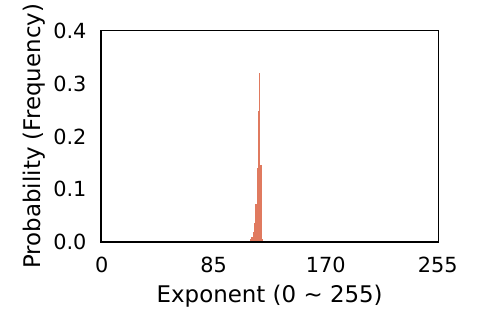} &
\includegraphics[width=0.485\linewidth]{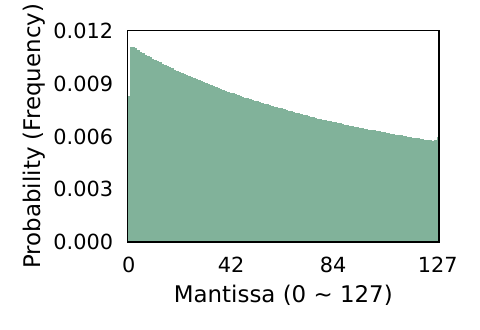}
\vspace{-3pt}\\
\multicolumn{2}{c}{\small (a) Experts of Layer 0 in Mixtral-8$\times$7B-Instruct \cite{jiang24}} \\
\includegraphics[width=0.485\linewidth]{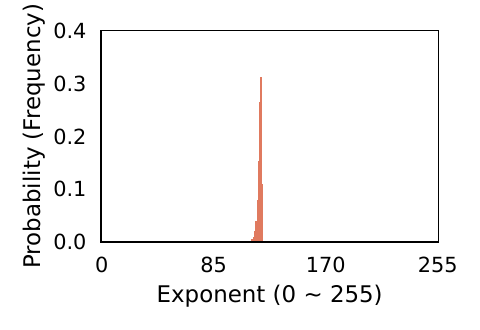} &
\includegraphics[width=0.485\linewidth]{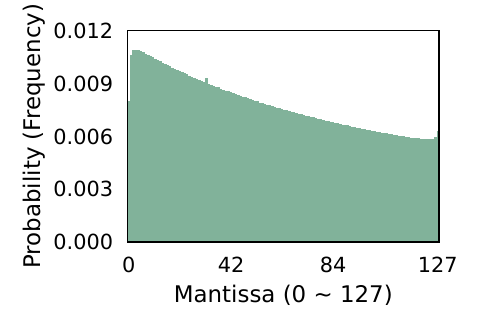}
\vspace{-3pt}\\
\multicolumn{2}{c}{\small (b) Experts of Layer 20 in Qwen3-Next-80B-A3B-Instruct \cite{qwen3technicalreport}}
\end{tabular}
\vspace{-6pt}
\captionof{figure}{Exponent and mantissa distribution analysis of MoE expert weights.}
\label{fig:parameter_distribution}
\vspace{-9pt}
\end{figure}

Figure~\ref{fig:parameter_distribution} presents
one representative layer from each model.
The exponent bits (8~bits in BFloat16) are highly concentrated within a narrow
range of magnitudes, exhibiting a sharp probability peak and correspondingly
low entropy. 
In contrast, the mantissa fields (7~bits) are nearly uniformly distributed,
suggesting limited compression gains from their empirical symbol frequencies.
This contrast in field-level entropy is consistent across the profiled models and layers.
Note that prior studies \cite{zhang25,hershcovitch25,heilper25,fan26} have also
identified the distinct entropy patterns in mantissas and exponents
in traditional dense models. Our profiling targets the sparse computing paradigm
in MoE and confirms that MoE expert weights show a highly compressible exponent structure.

\section{Additional Evaluation Results}
\label{sec:additional-evaluation}

\begin{figure*}[!t]
\centering
\includegraphics[height=15pt]{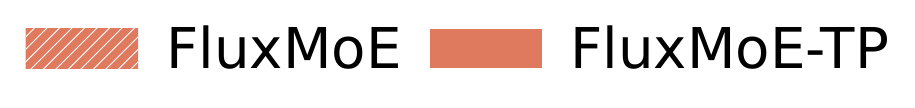}\par\vspace{-3pt}
\begin{tabular}{@{}c@{}c@{}c@{}c@{}}
\includegraphics[width=0.24\linewidth]{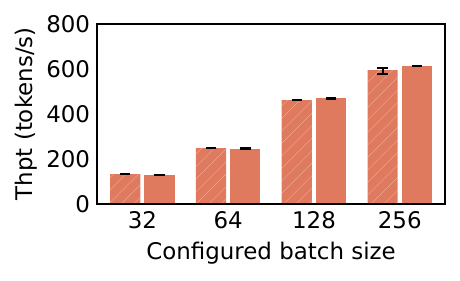} &
\includegraphics[width=0.24\linewidth]{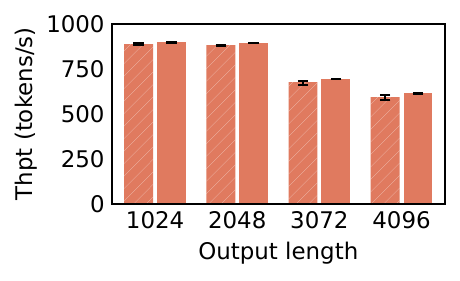} &
\includegraphics[width=0.24\linewidth]{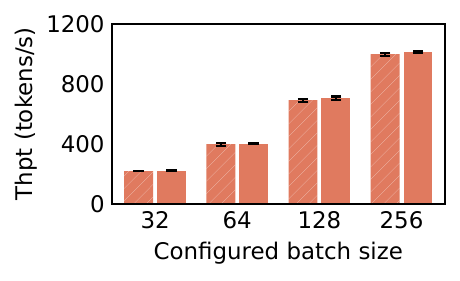} &
\includegraphics[width=0.24\linewidth]{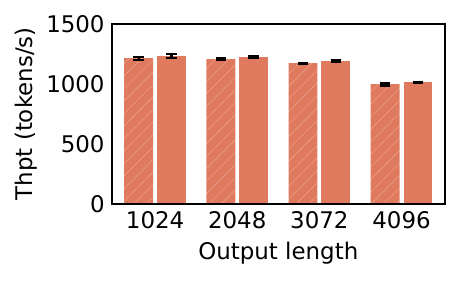} \\[-3pt]
\includegraphics[width=0.24\linewidth]{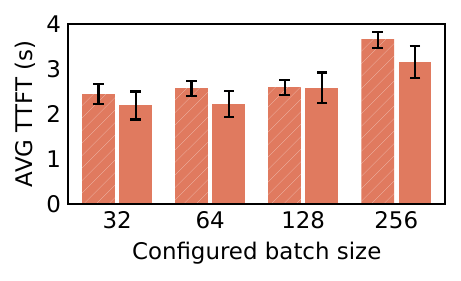} &
\includegraphics[width=0.24\linewidth]{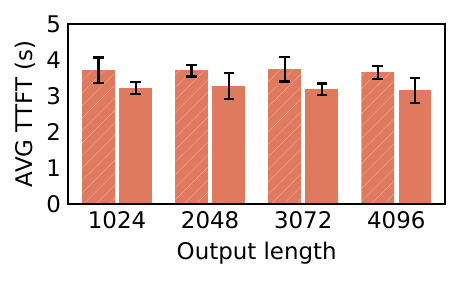} &
\includegraphics[width=0.24\linewidth]{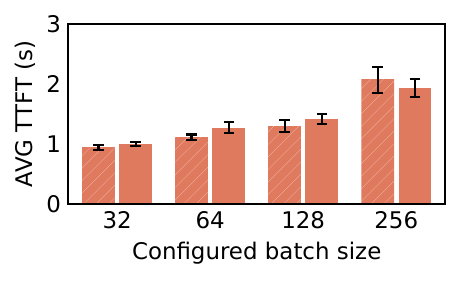} &
\includegraphics[width=0.24\linewidth]{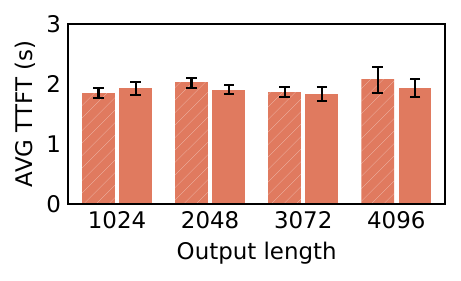} \\[-3pt]
\includegraphics[width=0.24\linewidth]{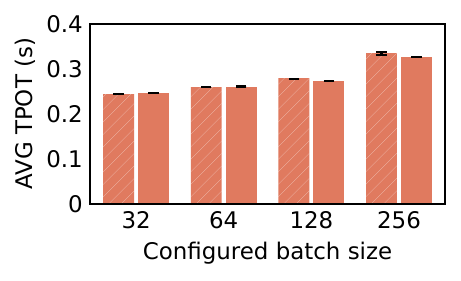} &
\includegraphics[width=0.24\linewidth]{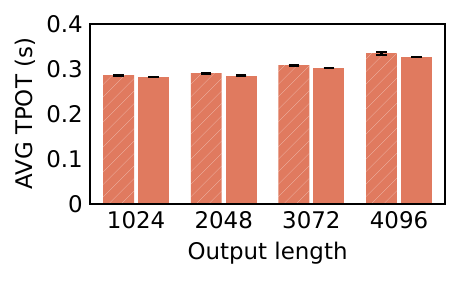} &
\includegraphics[width=0.24\linewidth]{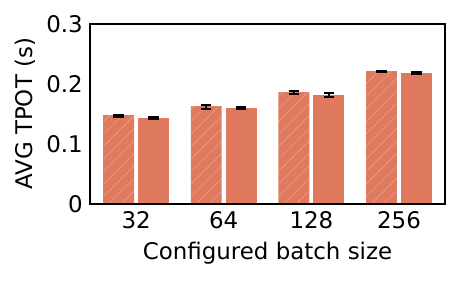} &
\includegraphics[width=0.24\linewidth]{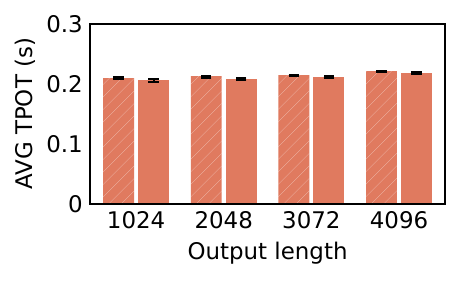} \\[-3pt]
\multicolumn{2}{c}{(a) GLM-4.5} &
\multicolumn{2}{c}{(b) Qwen3-Next}
\end{tabular}
\vspace{-6pt}
\caption{Offline performance of \sysname and \sysname-TP.}
\vspace{-6pt}
\label{fig:fluxmoe-tp-offline}
\end{figure*}

\begin{figure}[!t]
\ContinuedFloat
\centering
\includegraphics[height=15pt]{figs/legend_bar_fluxmoe_tp.pdf}\par\vspace{-3pt}
\begin{tabular}{@{}c@{}c@{}}
\includegraphics[width=0.49\linewidth]{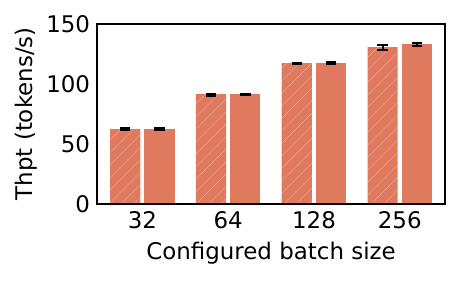} &
\includegraphics[width=0.49\linewidth]{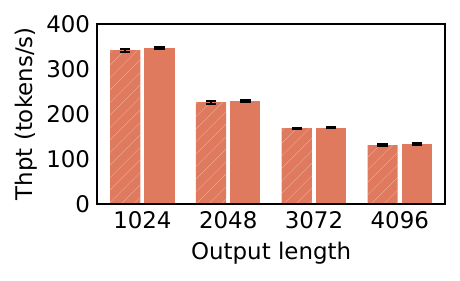} \\[-3pt]
\includegraphics[width=0.49\linewidth]{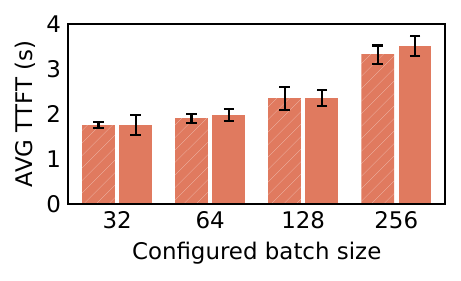} &
\includegraphics[width=0.49\linewidth]{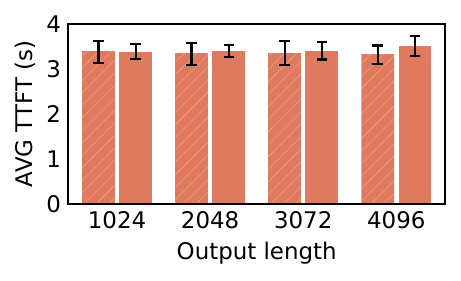} \\[-3pt]
\includegraphics[width=0.49\linewidth]{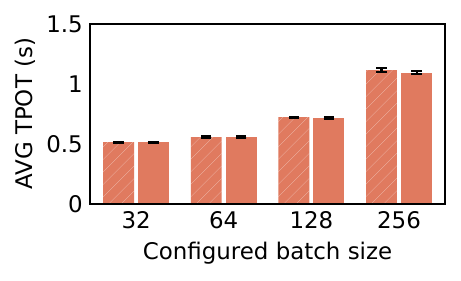} &
\includegraphics[width=0.49\linewidth]{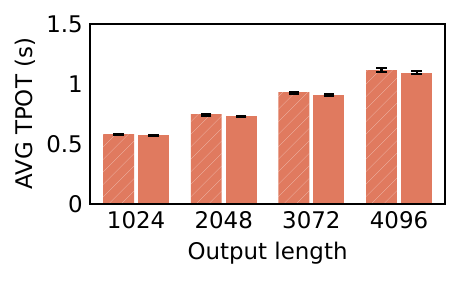}
\end{tabular}
\par
{\small (c) Mixtral}
\vspace{-6pt}
\caption{Offline performance of \sysname and \sysname-TP (continued).}
\vspace{-12pt}
\end{figure}

\begin{figure}[!t]
\centering
\includegraphics[height=15pt]{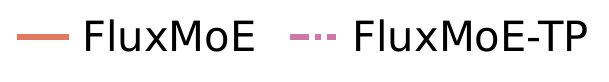}\par\vspace{-3pt}
\begin{tabular}{@{}c@{}c@{}c@{}}
\includegraphics[width=0.325\linewidth]{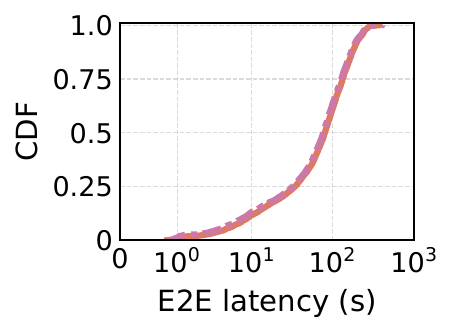} &
\includegraphics[width=0.325\linewidth]{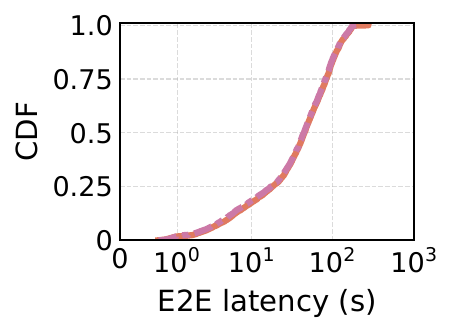} &
\includegraphics[width=0.325\linewidth]{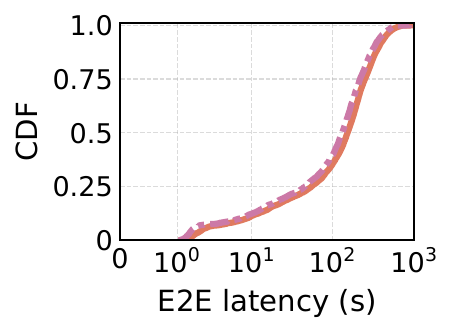} \\[-3pt]
\multicolumn{3}{c}{\small (a) Concurrency limit = 256} \\[1pt]
\includegraphics[width=0.325\linewidth]{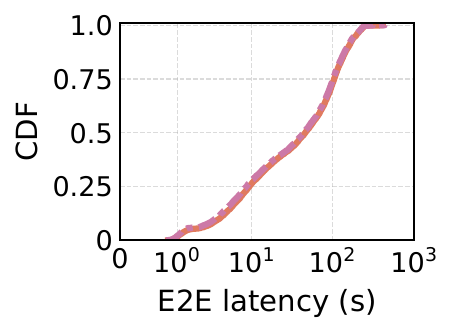} &
\includegraphics[width=0.325\linewidth]{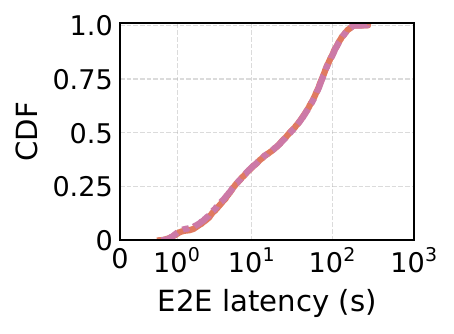} &
\includegraphics[width=0.325\linewidth]{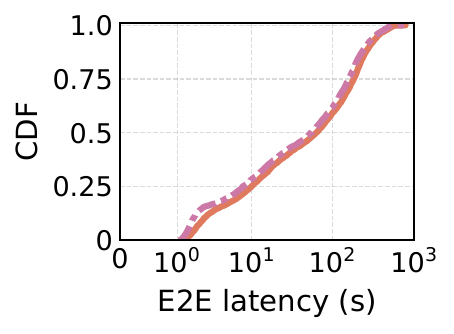} \\[-3pt]
\multicolumn{3}{c}{\small (b) Concurrency limit = 512}
\end{tabular}
\vspace{-6pt}
\caption{Online performance of \sysname and
\sysname-TP. From left to right: GLM-4.5,
Qwen3-Next, and
Mixtral.}
\label{fig:fluxmoe-tp-online}
\vspace{-6pt}
\end{figure}

We supplement \S\ref{sec:eval} with four sets of results: (i) offline and online
performance under TP-only \sysname execution (Figures~\ref{fig:fluxmoe-tp-offline} and \ref{fig:fluxmoe-tp-online});
(ii) intermediate-batch materialization
measurements for Exp\#4 (Table~\ref{tab:additional_materialization_breakdown});
(iii) planner-sensitivity data for Exp\#5 (Table~\ref{tab:exp5-settings}); and
(iv) online serving results at additional request rates for Exp\#7 (Figure~\ref{fig:exp7-additional-rates}).

\para{\sysname-TP offline and online performance.}
Because KTransformers \cite{chen25} does not support EP, we implement
a TP-only variant, \sysname-TP, retaining the mechanisms in
\S\ref{sec:design}. We evaluate it offline and online with the same GPU
deployments and residency configurations as \sysname, using TP degrees of 8,
4, and 2 for GLM-4.5, Qwen3-Next, and Mixtral, respectively.

\begin{figure*}[!t]
\centering
\small
\captionof{table}{(Exp\#4) Additional expert-materialization results
at $B_{\mathrm{cfg}}\in\{64,128\}$ and at output length 4,096
(mean $\pm$ std., ms). CPU Offloading and Decompression report isolated
PCIe-transfer and decompression time, respectively. E2E materialization spans
load issuance to tensor readiness, including queueing, stream waits, and launch
delay; Stall-free window spans issuance until tensor use; AVG stall is the
remaining wait, averaged across layers. For \sysname-H, C and D denote CPU-
and compressed-GPU-backed layers, respectively.}
\label{tab:additional_materialization_breakdown}
\vspace{-6pt}
\scriptsize
\setlength{\tabcolsep}{3.5pt}
\renewcommand{\arraystretch}{0.96}
\resizebox{0.97\textwidth}{!}{%
\begin{tabular}{@{}cccccccc@{}}
\toprule[1.2pt]
\textbf{Model} & \textbf{Method} & $\mathbf{B}_{\mathbf{cfg}}$ &
\textbf{CPU Offloading} & \textbf{Decompression} & \textbf{E2E materialization} &
\textbf{Stall-free window} & \textbf{AVG stall} \\
\midrule[0.8pt]
\multirow{2}{*}{GLM-4.5}
& \sysname & 64  & -- & $2.17\pm0.01$ & $2.17\pm0.01$ &
$5.37\pm0.54$ & $<0.01$ \\[2pt]
& \sysname & 128 & -- & $2.17\pm0.01$ & $2.17\pm0.01$ &
$5.68\pm0.90$ & $<0.01$ \\
\midrule[0.5pt]
\multirow{4}{*}[-1.5em]{Mixtral}
& \sysname   & 64  & $15.28\pm4.97$ & $3.13\pm0.15$ &
$27.84\pm1.66$ & $21.64\pm6.50$ & $7.17\pm4.76$ \\
& \sysname-H & 64  & $155.02\pm16.82$ & $3.51\pm0.02$ &
\makecell{C: $207.96\pm27.75$\\D: $3.51\pm0.02$} &
\makecell{C: $106.50\pm75.59$\\D: $12.65\pm13.77$} &
$13.00\pm39.19$ \\
\cmidrule(l){2-8}
& \sysname   & 128 & $15.30\pm4.91$ & $3.13\pm0.15$ &
$28.08\pm1.49$ & $20.86\pm6.68$ & $7.85\pm4.93$ \\
& \sysname-H & 128 & $146.10\pm13.41$ & $3.51\pm0.01$ &
\makecell{C: $206.70\pm29.06$\\D: $3.52\pm0.02$} &
\makecell{C: $106.37\pm72.60$\\D: $13.31\pm15.69$} &
$13.05\pm38.17$ \\
\bottomrule[1.2pt]
\end{tabular}%
}
\vspace{-3pt}
\end{figure*}

\begin{figure*}[!t]
\centering
\small
\captionof{table}{(Exp\#5) Released down-projection tensors and GPU memory
across all GPUs. Each cell reports released storage / total compressed down-projection storage;
fused gate/up-projection tensors remain untouched.}
\label{tab:exp5-settings}
\vspace{-6pt}
\setlength{\tabcolsep}{3.5pt}
\renewcommand{\arraystretch}{1.05}
\begin{tabular}{@{}cc*{6}{c}@{}}
\toprule[1.2pt]
\multicolumn{2}{c}{\multirow{2}{*}[-7pt]{\textbf{Setting}}} &
\multicolumn{2}{c}{\textbf{GLM-4.5}} &
\multicolumn{2}{c}{\textbf{Qwen3-Next}} &
\multicolumn{2}{c}{\textbf{Mixtral}} \\
\cmidrule(lr){3-4}\cmidrule(lr){5-6}\cmidrule(lr){7-8}
\multicolumn{2}{c}{} &
\makecell{\textbf{Tensors}\\\textbf{(released/total)}} &
\makecell{\textbf{Memory (GB)}\\\textbf{(released/total)}} &
\makecell{\textbf{Tensors}\\\textbf{(released/total)}} &
\makecell{\textbf{Memory (GB)}\\\textbf{(released/total)}} &
\makecell{\textbf{Tensors}\\\textbf{(released/total)}} &
\makecell{\textbf{Memory (GB)}\\\textbf{(released/total)}} \\
\midrule[0.8pt]
\multicolumn{2}{c}{All-compressed} & $0/14{,}240$ & $0/151.2$ & $0/24{,}576$ &
$0/35.0$ & $0/256$ & $0/20.3$ \\
\midrule[0.5pt]
AG=0.25 & CI=8 & $9{,}200/14{,}240$ & $97.7/151.2$ &
$9{,}216/24{,}576$ & $13.1/35.0$ & $42/256$ & $3.3/20.3$ \\
\textbf{AG=0.25} & \textbf{CI=16} & $4{,}600/14{,}240$ & $48.8/151.2$ &
$4{,}864/24{,}576$ & $6.9/35.0$ & $24/256$ & $1.9/20.3$ \\
AG=0.25 & CI=32 & $2{,}320/14{,}240$ & $24.6/151.2$ &
$2{,}432/24{,}576$ & $3.5/35.0$ & $18/256$ & $1.4/20.3$ \\
AG=0.25 & CI=64 & $1{,}200/14{,}240$ & $12.7/151.2$ &
$1{,}280/24{,}576$ & $1.8/35.0$ & $14/256$ & $1.1/20.3$ \\
\midrule[0.5pt]
\textbf{CI=16} & \textbf{AG=0.25} & $4{,}600/14{,}240$ & $48.8/151.2$ &
$4{,}864/24{,}576$ & $6.9/35.0$ & $24/256$ & $1.9/20.3$ \\
CI=16 & AG=0.50 & $9{,}200/14{,}240$ & $97.7/151.2$ &
$9{,}216/24{,}576$ & $13.1/35.0$ & $40/256$ & $3.2/20.3$ \\
CI=16 & AG=0.75 & $9{,}960/14{,}240$ & $105.7/151.2$ &
$9{,}600/24{,}576$ & $13.7/35.0$ & $54/256$ & $4.3/20.3$ \\
CI=16 & AG=1.00 & $11{,}680/14{,}240$ & $124.0/151.2$ &
$9{,}728/24{,}576$ & $13.8/35.0$ & $72/256$ & $5.7/20.3$ \\
\bottomrule[1.2pt]
\end{tabular}
\vspace{-3pt}
\end{figure*}

We reuse \sysname's offline results from Exp\#1 and Exp\#2
(Figures~\ref{fig:exp1} and~\ref{fig:exp2}) and its online serving results from Exp\#7
(Figure~\ref{fig:exp7}). Figures~\ref{fig:fluxmoe-tp-offline}
and~\ref{fig:fluxmoe-tp-online} compare them with \sysname-TP.
In offline evaluations, \sysname-TP stays within 3.9\% of \sysname in
throughput and 2.6\% in average TPOT. Across all evaluated model-workload
settings, \sysname-TP reduces \sysname's mean TTFT by at most 14.9\%.
For online serving evaluations, \sysname-TP's
P50 and P95 E2E latencies differ from \sysname's by at most 4.8\% and 3.8\%,
respectively, on GLM-4.5 and Qwen3-Next.
On Mixtral, \sysname-TP reduces P50 by 13.6\% (14.4\%) and P95 by
12.0\% (6.7\%) at concurrency limit 256 (512).
The similar offline and online performance of \sysname and \sysname-TP shows that
\sysname's gains over KTransformers do not come from enabling EP.

\para{(Exp\#4) Materialization across batch sizes.}
Table~\ref{tab:additional_materialization_breakdown} extends the per-layer
materialization breakdown in Table~\ref{tab:materialization_breakdown} to the
intermediate $B_{\mathrm{cfg}}=64$ and 128 settings, omitted from the main
paper for space.
At both intermediate batches, GLM-4.5 exposes less than 0.01\,ms of average
stall, consistent with the $B_{\mathrm{cfg}}=32$ and 256 results shown in
Table~\ref{tab:materialization_breakdown}. Across all four batches, GLM-4.5's
E2E materialization remains 2.17\,ms, while Mixtral's varies by at most 1.1\%,
showing that increasing batch size does not systematically slow
materialization. At $B_{\mathrm{cfg}}=64$ and 128, \sysname reduces Mixtral's
average stall by 39.8--44.8\% relative to \sysname-H, consistent with the stall
reductions at $B_{\mathrm{cfg}}=32$ and 256 in
Table~\ref{tab:materialization_breakdown}.

\para{(Exp\#5) Tensor release across CI and AG settings.}
Table~\ref{tab:exp5-settings} reports released tensor counts and storage
volumes for all CI and AG settings, extending the default CI=16 and AG=0.25
results in \S\ref{subsec:planner_sensitivity}.
All runs stop before exhausting the down-projection tensors, so these results
cover only the planner's first tensor-level stage; no fused gate/up-projection
tensor is offloaded.
At CI=16 and AG=0.25, the planner releases 4,600, 4,864, and 24 down-projection
tensors for GLM-4.5, Qwen3-Next, and Mixtral, respectively.
This frees 48.8\,GB, 6.9\,GB, and 1.9\,GB of GPU memory, equivalent to
32.3\%, 19.7\%, and 9.4\% of their compressed down-projection storage.
Relative to each model's total compressed routed-expert storage, the released
fractions are 10.8\%, 6.6\%, and 3.1\%, respectively.
The released fractions differ across models since the planner stops
offloading when additional loading would delay inference, rather than
always targeting a fixed release fraction.
GLM-4.5 permits more adjustments since H20 provides higher PCIe
bandwidth.
For Mixtral, $L_r=4$, so each of the 12 AG=0.25 adjustments offloads one
down-projection tensor per GPU rank, releasing 24 tensors across two ranks.

At AG=0.25, increasing CI from 8 to 64 reduces released GPU memory by 87.0\%,
86.3\%, and 66.7\% for GLM-4.5, Qwen3-Next, and Mixtral, respectively.
Released GPU memory decreases across all three models since a larger CI
invokes the planner less often, allowing fewer reclamation steps before the
workload completes.

\begin{figure*}[!t]
\centering
\includegraphics[height=15pt]{figs/legend_line_all.pdf}\par\vspace{-3pt}
\begin{tikzpicture}
\node[inner sep=0pt] (requestRatePanels) {%
\begin{tabular}{@{}c@{}c@{}c@{\hspace{4pt}}c@{}c@{}c@{}}
\includegraphics[width=0.16\linewidth]{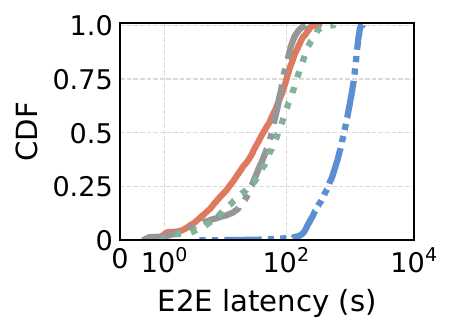} &
\includegraphics[width=0.16\linewidth]{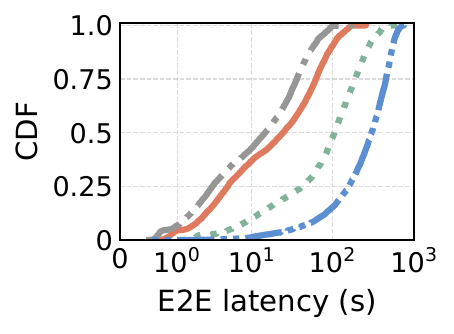} &
\includegraphics[width=0.16\linewidth]{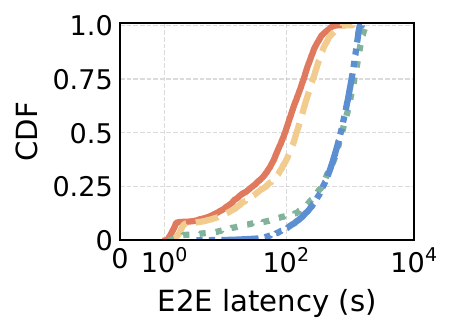} &
\includegraphics[width=0.16\linewidth]{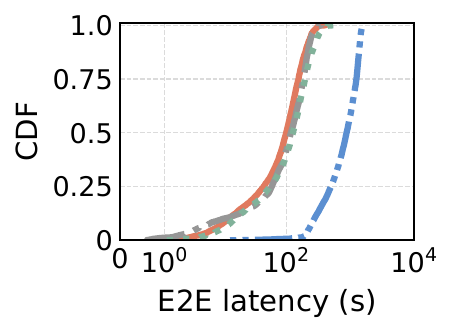} &
\includegraphics[width=0.16\linewidth]{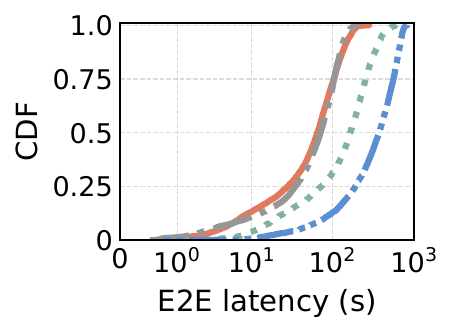} &
\includegraphics[width=0.16\linewidth]{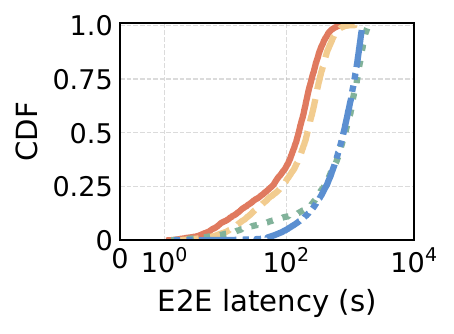} \\[-3pt]
\multicolumn{3}{c}{\small (a) Concurrency limit = 256} &
\multicolumn{3}{c}{\small (a) Concurrency limit = 256} \\[1pt]
\includegraphics[width=0.16\linewidth]{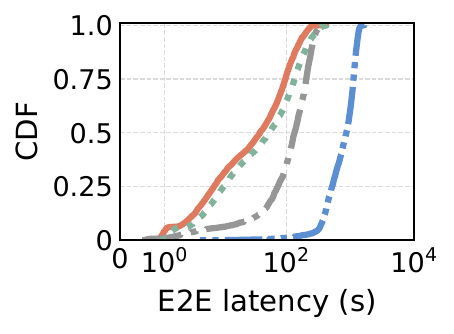} &
\includegraphics[width=0.16\linewidth]{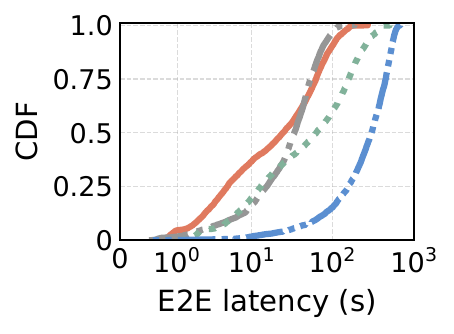} &
\includegraphics[width=0.16\linewidth]{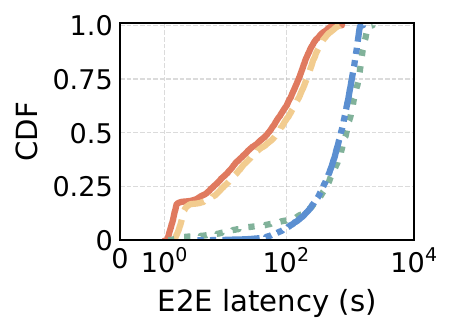} &
\includegraphics[width=0.16\linewidth]{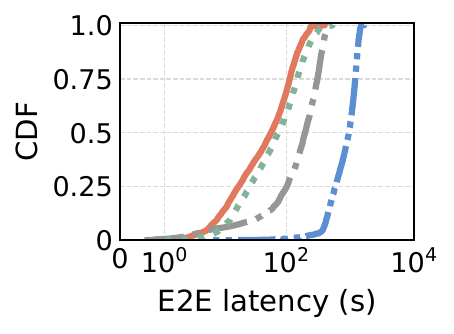} &
\includegraphics[width=0.16\linewidth]{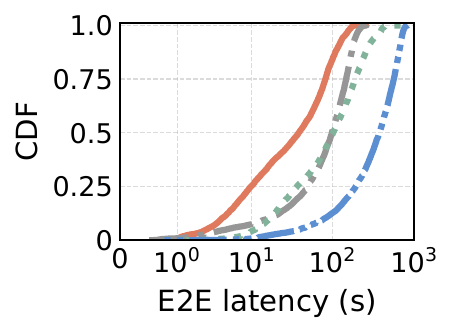} &
\includegraphics[width=0.16\linewidth]{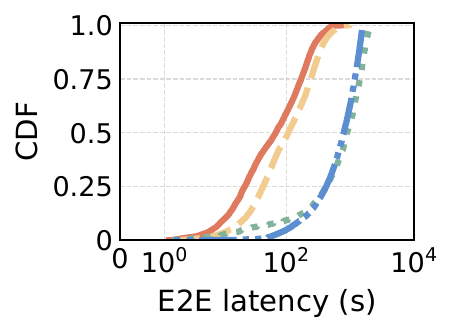} \\[-3pt]
\multicolumn{3}{c}{\small (b) Concurrency limit = 512} &
\multicolumn{3}{c}{\small (b) Concurrency limit = 512}
\end{tabular}%
};
\draw[black!70, line width=1pt, dash pattern=on 4pt off 1.8pt]
(requestRatePanels.north) -- (requestRatePanels.south);
\end{tikzpicture}
\vspace{-6pt}
\caption{(Exp\#7) Online performance at 5 requests/s (left) and
20 requests/s (right). For each panel, from left to right: GLM-4.5, Qwen3-Next, and Mixtral.}
\label{fig:exp7-additional-rates}
\end{figure*}

\begin{figure}[!t]
\ContinuedFloat
\centering
\includegraphics[width=\columnwidth]{figs/legend_line_all.pdf}\par\vspace{-3pt}
\begin{tabular}{@{}c@{}c@{}c@{}}
\includegraphics[width=0.325\linewidth]{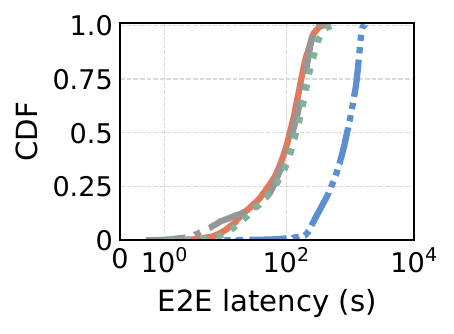} &
\includegraphics[width=0.325\linewidth]{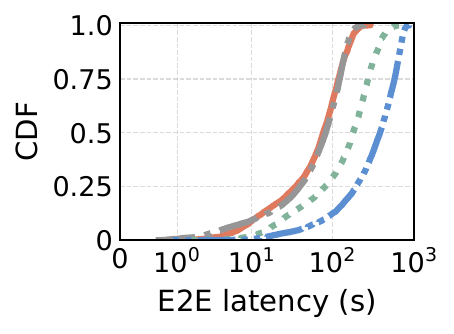} &
\includegraphics[width=0.325\linewidth]{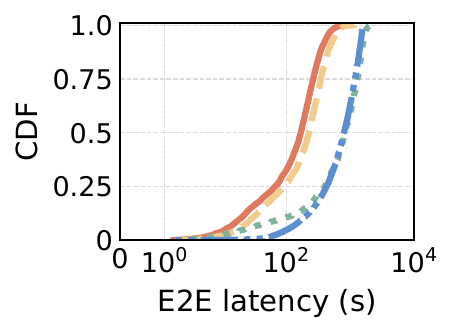} \\[-3pt]
\multicolumn{3}{c}{\small (a) Concurrency limit = 256} \\[1pt]
\includegraphics[width=0.325\linewidth]{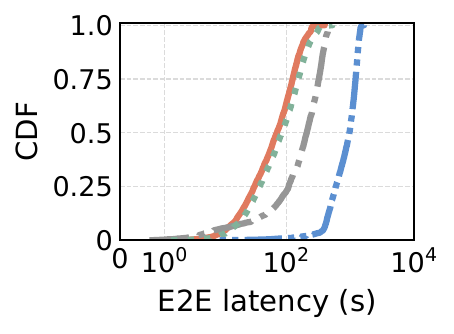} &
\includegraphics[width=0.325\linewidth]{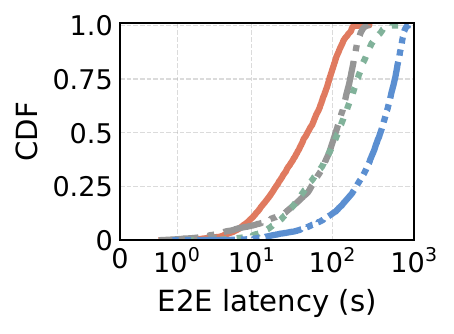} &
\includegraphics[width=0.325\linewidth]{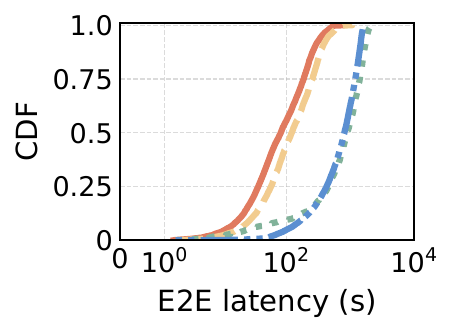} \\[-3pt]
\multicolumn{3}{c}{\small (b) Concurrency limit = 512}
\end{tabular}
\vspace{-6pt}
\caption{(Exp\#7) Online performance at 40 requests/s.
From left to right: GLM-4.5, Qwen3-Next, and Mixtral.}
\vspace{-9pt}
\end{figure}

At CI=16, increasing AG from 0.25 to 1.0 raises the released fraction of
compressed down-projection storage from 32.3\% to 82.0\% for GLM-4.5, 19.7\%
to 39.4\% for Qwen3-Next, and 9.4\% to 28.1\% for Mixtral.
The total released tensor count does not scale linearly with AG: a larger AG
offloads more tensors per reclamation but also increases materialization demand,
causing the planner to trigger fewer reclamations before completion.
For GLM-4.5 and Qwen3-Next, increasing AG from 0.25 to 0.5 doubles the
per-step tensor count. Since the number of adjustments changes little, the
total released tensor count nearly doubles.
Above AG=0.5, larger per-step releases increase materialization time and
cause the planner to stop offloading earlier, so total release grows more slowly.
Mixtral's released fraction grows nearly linearly with AG. Each compressed
tensor in Mixtral is large, so increasing AG substantially increases the memory released
per adjustment, while the number of adjustments decreases only from 12 at
AG=0.25 to 9 at AG=1.0.

\para{(Exp\#7) Online serving at additional request rates.}
We report all three models at 10 requests/s in \S\ref{subsec:online_performance};
Figure~\ref{fig:exp7-additional-rates}
extends the evaluation to 5, 20, and 40 requests/s to cover lighter and heavier
loads. At 5 requests/s, \sysname on GLM-4.5 is saturated at a concurrency
limit of 256 but unsaturated at 512. On Qwen3-Next, both \sysname and vLLM are
unsaturated under both limits. All remaining settings are saturated.

At 5 requests/s, \sysname's P95 latency on Qwen3-Next is 66.0\% and 30.5\%
higher than vLLM's at limits of 256 and 512, respectively. In these runs,
$B_{\mathrm{act}}$ is limited by ready requests rather than KV capacity, so
\sysname gains no batching benefit to offset its materialization overhead.
For GLM-4.5 at a concurrency limit of 256, \sysname's P95 is 39.0\%
higher than vLLM's.
The reason is that the 256 limit prevents \sysname
from using its additional KV capacity to increase $\overline{B}_{\mathrm{act}}$, but it still
incurs per-layer materialization overhead.
At a limit of 512, \sysname's P95 is 33.7\% lower than vLLM's.
Raising the limit from 256 to 512 allows \sysname to exploit its additional KV
capacity, sustain a larger $\overline{B}_{\mathrm{act}}$ than vLLM, and reduce
queueing.
For Mixtral, \sysname lowers P95 relative to \sysname-H by 27.5\% and 22.4\%
at limits of 256 and 512, respectively, since bandwidth-balanced backends
reduce materialization overhead under the same offload budget.

At 20 and 40 requests/s, every system saturates for all three models under both
concurrency limits. \sysname achieves the lowest P50 in every case. 
For instance, at a concurrency limit of 256 and 20 requests/s, \sysname reduces
P50 by 15.5\% over vLLM on GLM-4.5, 8.7\% over vLLM on Qwen3-Next, and
26.8\% over \sysname-H on Mixtral.
At 40 requests/s, the corresponding reductions are 8.6\%, 6.8\%, and 25.5\%, respectively.
For GLM-4.5 and Qwen3-Next, additional KV capacity supports larger $B_{\mathrm{act}}$
than vLLM. For Mixtral, \sysname's gains over \sysname-H instead reflect lower
materialization overhead at the same KV capacity.

Overall, these additional request rates contextualize the 10 requests/s results
in \S\ref{subsec:online_performance}: extra KV capacity benefits serving when
it can increase $B_{\mathrm{act}}$; limited request availability or a
restrictive concurrency limit can leave that capacity unused
while materialization overhead remains.

\begin{figure*}[!t]
\centering
\small
\captionof{table}{(Exp\#8) Perplexity evaluation of the three models on the
WikiText-2-Raw-v1 \cite{merity17} test split.}
\label{tab:ppl-evaluation-details}
\vspace{-6pt}
\setlength{\tabcolsep}{6pt}
\renewcommand{\arraystretch}{1.3}
\begin{tabular*}{\textwidth}{@{\extracolsep{\fill}}cccccccc@{}}
\toprule[1.2pt]
\multirow{2}{*}{\textbf{Model}} &
\multirow{2}{*}{\textbf{Tokenizer}} &
\multirow{2}{*}{\makecell{\textbf{Scored}\\\textbf{tokens}}} &
\multirow{2}{*}{\makecell{\textbf{Number of}\\\textbf{windows}}} &
\multirow{2}{*}{\textbf{TP/EP}} &
\multicolumn{3}{c}{\textbf{Perplexity}} \\
\cmidrule(lr){6-8}
& & & & & \textbf{vLLM} & \textbf{\sysname} &
\boldmath$\lvert\Delta\rvert$ \\
\midrule[0.8pt]
GLM-4.5 &
\texttt{PreTrainedTokenizerFast} &
289,722 &
563 &
8/8 & 3.6497 & 3.6497 & 0.0000 \\
Qwen3-Next &
\texttt{Qwen2Tokenizer} &
299,077 &
582 &
4/4 & 5.2456 & 5.2456 & 0.0000 \\
Mixtral &
\texttt{LlamaTokenizerFast} &
334,659 &
651 &
4/4 & 3.6976 & 3.6976 & 0.0000 \\
\bottomrule[1.2pt]
\end{tabular*}
\end{figure*}

\section{Model-Quality Evaluation}
\label{sec:additional-exp}

We evaluate whether \sysname's dynamic materialization and lossless
compression affect model quality.

\para{(Exp\#8) Perplexity evaluation.}
Following prior work on LLM compression \cite{frantar23,lin24}, we use
perplexity on the complete test split of WikiText-2-Raw-v1 \cite{merity17} to
evaluate whether \sysname preserves the model quality of
weight-resident vLLM \cite{kwon23}.  We evaluate the same three MoE models used
in our performance experiments: GLM-4.5, Qwen3-Next, and Mixtral.
For each model, both systems
use the same checkpoint, tokenizer, TP/EP configuration, and evaluation inputs.
For Mixtral, we use TP/EP degrees of
4/4 since weight-resident vLLM \cite{kwon23} runs out of GPU memory at a TP
degree of 2 on L40S; TP=4 allows both systems to complete the perplexity
comparison using the same checkpoint.
To construct the evaluation input, we concatenate all text entries in the test
split using double-newline separators and tokenize the resulting text as a
single continuous sequence with each model's native tokenizer.  We do not add
beginning-of-sequence (BOS) or end-of-sequence (EOS) tokens.  We use a sliding context window of 2,048 tokens with a
stride of 512 tokens.  Overlapping tokens provide context only; for each window,
we score only the newly introduced tokens.
We process each window independently with a batch size of one, without cross-window KV
cache reuse or prefix caching.

Table~\ref{tab:ppl-evaluation-details} compares the perplexity obtained by
vLLM and \sysname. Since each model uses its native
tokenizer, the same WikiText-2 input produces different numbers of scored tokens
and sliding windows. vLLM and \sysname achieve identical perplexities for
all three models: 3.6497 for GLM-4.5, 5.2456 for
Qwen3-Next, and 3.6976 for
Mixtral,
demonstrating that \sysname's dynamic expert
materialization and lossless decompression introduce no measurable model-quality
loss relative to vLLM at the reported precision.

\end{document}